\documentclass[11pt,a4paper]{article}
\usepackage[hyperref]{acl2020}
\usepackage{times}
\usepackage{latexsym}

\usepackage{tabularx}
\usepackage{footnote}
\usepackage{graphicx}
\usepackage{multirow}
\usepackage{amssymb}
\usepackage{amsmath}
\usepackage{subfig}

\usepackage{microtype}

\aclfinalcopy 

\title{Estimating predictive uncertainty for rumour verification models}

\author{Elena Kochkina \\
  University of Warwick, UK \\
  The Alan Turing Institute, UK \\
  \texttt{E.Kochkina@warwick.ac.uk} \\\And
  Maria Liakata \\
  Queen Mary University of London, UK \\
  The Alan Turing Institute, UK \\
  University of Warwick, UK \\
  \texttt{mliakata@turing.ac.uk} \\}

\date{}
\begin{document}
\maketitle
\begin{abstract}
 The inability to correctly resolve rumours circulating online can have harmful real-world consequences. 
 We present a method for incorporating model and data uncertainty estimates into natural language processing models for automatic rumour verification.  We show that these estimates can be used to filter out model predictions likely to be erroneous, so that these difficult instances can be prioritised by a human fact-checker. We propose two methods for uncertainty-based instance rejection, supervised and unsupervised. We also show how uncertainty estimates can be used to interpret model performance as a rumour unfolds. 
\end{abstract}
\section{Introduction}
One of the greatest challenges of the information age is the rise of pervasive misinformation. Social media platforms enable it to spread rapidly, reaching wide audiences before manual verification can be performed. Hence there is a strive to create automated tools that assist with rumour resolution. 
Information about unfolding real-world events such as natural disasters often appears in a piece-wise manner, making verification a time-sensitive problem.
Failure to identify misinformation can have a harmful impact, thus it is desirable that an automated system aiding rumour verification does not only make a judgement but that it can also inform a human fact-checker of its uncertainty.

Deep learning models are currently the state-of-the-art in many Natural Language Processing (NLP) tasks, including rumour detection \citep{ma2018rumor}, the task of identifying candidate rumours, and rumour verification \citep{li2019rumor,zhang2019reply}, where the goal is to resolve the veracity of a rumour. Latent features and large parameter spaces of deep learning models make it hard to interpret a model's decisions. Increasingly researchers are investigating methods for understanding model predictions, such as through analysing neural attention \citep{vaswani2017attention} and studying adversarial examples~\citep{yuan2019adversarial}. Another way to gain insights into a model's decisions is via estimating its uncertainty. Understanding what a model does not know can help us determine when we can trust its output and at which stage information needs to be passed on to a human~\citep{kendall2017uncertainties}. 

In this paper, rather than purely focusing on the performance of a rumour verification model, we estimate its predictive uncertainty to gain understanding of a model's decisions and filter out the cases that are 'hard' for the model. 
We consider two types of predictive uncertainty: data uncertainty (aleatoric) and model uncertainty (epistemic). The approach we adopt requires minimal changes to a given model and is relatively computationally inexpensive, thus making it possible to apply to various architectures. 

We make the following contributions:
\begin{itemize}
    \itemsep0em 
    \item We are the first to apply methods for uncertainty estimation to the problem of rumour verification. We show that removing instances with high uncertainty filters out many incorrect predictions, gaining performance improvement in the rest of the dataset.
    \item We propose a supervised method for instance removal that combines both aleatoric and epistemic uncertainty and outperforms an unsupervised approach. 
    \item We propose a way to analyse uncertainty patterns as a rumour unfolds in time. We make use of this to study the relation between the stance expressed in response tweets and fluctuation in uncertainty at the time step following a response.
    \item We explore the relationship between uncertainty estimates and class labels.
\end{itemize}
\section{Related Work}
\subsection{Rumour Verification}
 A rumour is a \textit{circulating story of questionable veracity, which is apparently credible but hard to verify, and produces sufficient skepticism/anxiety so as to motivate finding out the actual truth}~\citep{zubiaga2018detection}. Rumour detection and verification in online conversations have gained popularity as tasks in recent years~\citep{zubiaga2016analysing,ma2016detecting,enayet2017niletmrg}. 
Existing works aim to improve performance of supervised learning algorithms that classify claims, leveraging linguistic cues, network- and user-related features, propagation patterns, support among responses and conversation structure~\citep{derczynski2017semeval,gorrell2018rumoureval}.
Due to the nature of the task, each rumour can be considered as a new domain and existing models struggle with generalisability. 
Here we employ model-agnostic methods of uncertainty estimation that can provide performance improvements and insight on the working of the models to inspire further development. 

\subsection{Related Work on Uncertainty Estimation}
There is a growing body of literature which aims to estimate predictive uncertainty of deep neural networks (DNNs)~\citep{gal2016dropout,lakshminarayanan2017simple,malinin2018predictive}. 
\citet{gal2016dropout} have shown that application of Monte-Carlo (MC) Dropout at testing time can be used to derive an uncertainty estimate for a DNN. 
\citet{lakshminarayanan2017simple} estimate model uncertainty by using a set of predictions from an ensemble of DNNs, while \citet{malinin2018predictive} propose a specialised framework, Prior Networks, for modelling predictive uncertainty.
Here we focus on the dropout method proposed by \citet{gal2016dropout} as it is computationally inexpensive, relatively simple and does not interfere with model training. 

Within NLP \citet{xiao2018quantifying} have used aleatoric \citep{kendall2017uncertainties} and epistemic \citep{gal2016dropout} uncertainty estimates for Sentiment analysis and Named Entity Recognition.
\citet{dong2018confidence} used a modification of \citet{gal2016dropout} method to output confidence scores for Neural Semantic Parsing.

Rumour Verification is a task where levels of certainty play a crucial role because of the potentially high impact of erroneous decisions.
Moreover, unlike other tasks, it is a time-sensitive problem: as new information comes to light the level of certainty is expected to change giving insights into a model’s predictions. We therefore explore the dynamics of uncertainty as a discussion unfolds in section 6.3. 
Note that data and model uncertainty should not be confused with uncertainty expressed by a user in a post. Automatically identifying levels of uncertainty expressed in text is a challenging NLP task \citep{jean2016uncertainty,vincze2015uncertainty}, which could be complementary to predictive uncertainty in the case of rumour verification.\\
\textbf{Active Learning and Uncertainty:} Uncertainty estimates could be used in an Active Learning (AL) setup. 
This would involve using uncertainty estimates over the model's predictions to select instances whose manual labelling and addition to the training set would yield the most benefit ~\citep{olsson2009literature}.
 Active learning has been applied to various NLP tasks in the past~\citep{settles2008analysis}.
More recently \citet{siddhant2018deep} have shown that Bayesian active learning by disagreement, using uncertainty estimates provided either by Dropout~\citep{gal2016dropout} or Bayes-by-Backprop~\citep{blundell2015weight} significantly improves over i.i.d. baselines and usually outperforms classic uncertainty sampling on a number of NLP tasks and datasets.
\citet{bhattacharjee2017active,bhattacharjee2019identifying} applied AL to identifying misinformation in news and social media. 
Our work could be applied in an AL setup to close the loop in incrementally training a model for misinformation using predictive uncertainty.
\begin{figure}[]
    \centering
    \includegraphics[width=\columnwidth]{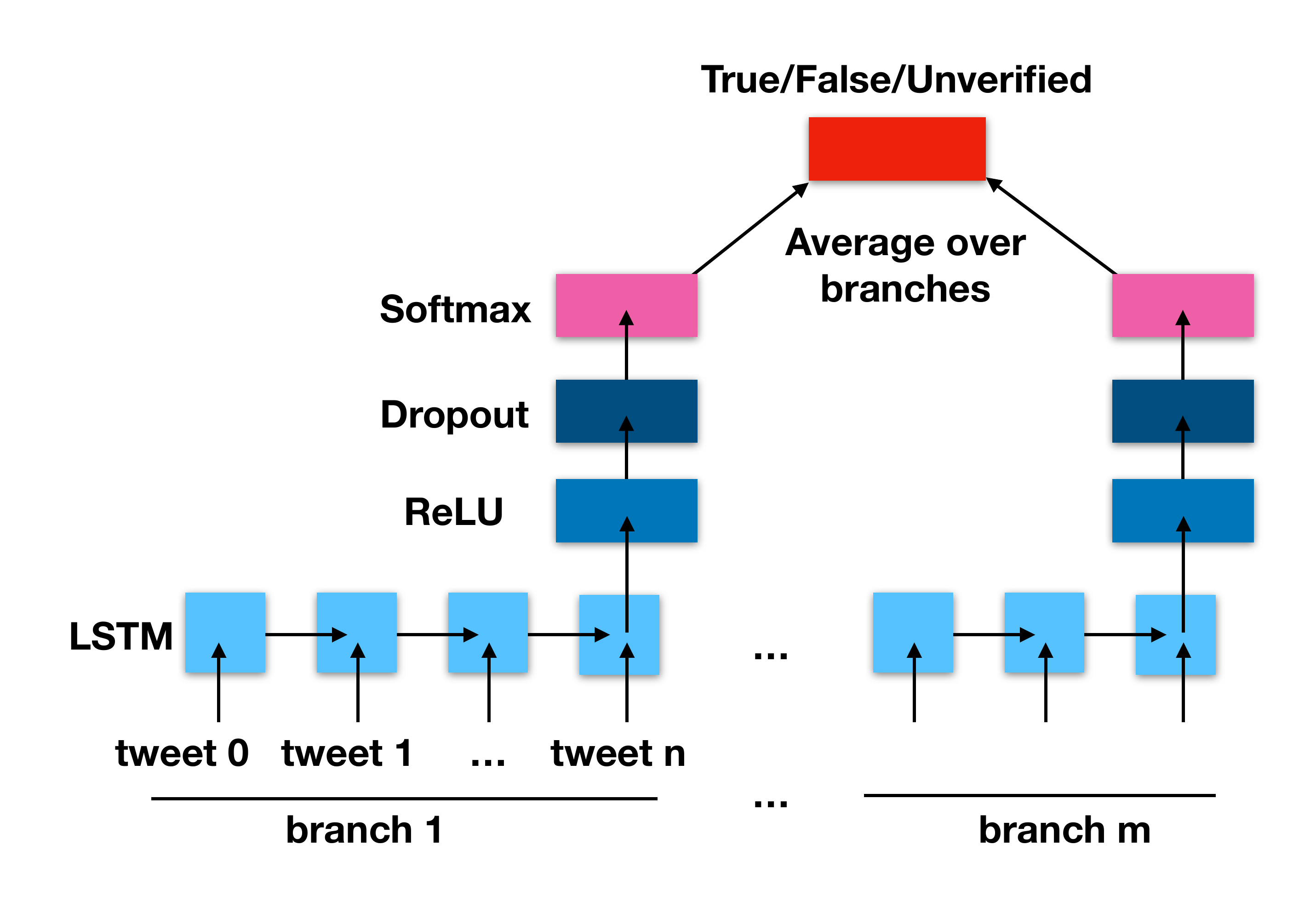}
    \caption{branch-LSTM model}
    \label{fig_model}
\end{figure}
\begin{figure*}
    \centering
    \includegraphics[width=\textwidth]{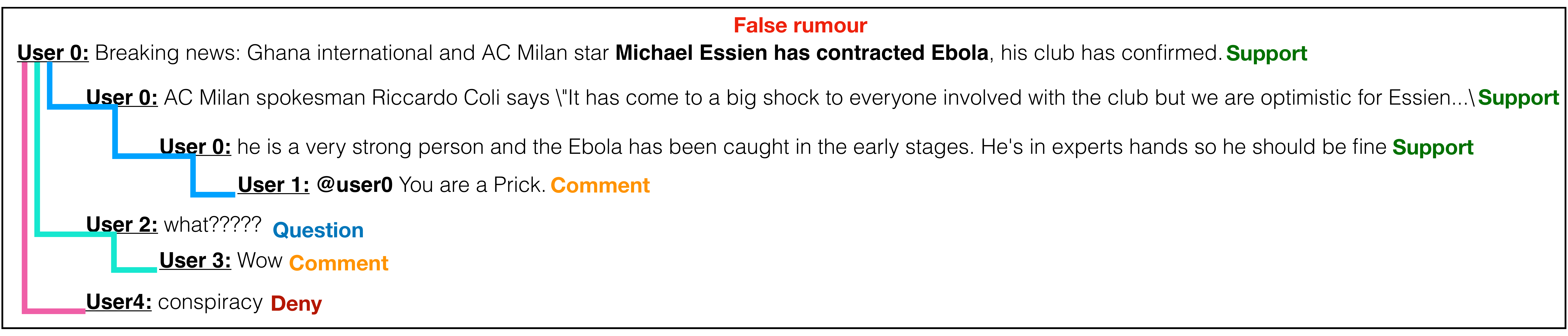}
    \caption{Example of a conversation from the PHEME dataset. Branches are highlighted as lines connecting the tweets.}
    \label{fig_conversation}
\end{figure*}
\section{Methodology}
\subsection{Rumour Verification Model}
We describe the rumour verification model which forms the basis of our experiments. This served as a competitive baseline model (branch-LSTM) for a Semeval task on rumour verification (RumourEval 2019)~\cite{gorrell2018rumoureval} \footnote{\url{https://github.com/kochkinaelena/RumourEval2019}}.
To process a conversation discussing a rumour while preserving some of the structural relations between the tweets, a tree-like conversation is split into branches, i.e linear sequences of tweets, as shown in Figure \ref{fig_conversation}. Branches are then used as training instances for a branch-LSTM model consisting of an LSTM layer followed by several ReLU layers and a softmax layer (default base of e and temperature of 1) that predicts class probabilities. Here we use outputs from the final time steps (see Figure~\ref{fig_model}). 
Given a training instance, branch of tweets $x_i$, $i \in [1,..,N]$, where $N$ is the number of branches, and the label $y_i$, represented as one-hot vector of size $C$, where $C$ is the number of classes, the loss function $l_1$ (categorical cross entropy) is calculated as follows: 
\begin{align*}
& u_i = f(x_i)\\
& v_i = W_vu_i + b_v\\
& p_i = softmax(v_i) = \frac{e^{v_i}}{\sum\limits_{k=1}^{C}e^{v_i^k}}\\
& l_1 = -\frac{1}{N}\sum\limits_{n=1}^{N}\sum\limits_{k=1}^{C}y_i^k log p_i^k,
\end{align*}
where $u_i$ is an intermediate output of layers prior to the softmax layer, $v_i$ is logits, and $p_i$ are predicted class probabilities for a training instance $x_i$. 
To obtain predictions for each of the conversation trees we average class probabilities for each of the branches in the tree.  In this case tweets are represented as the average of the corresponding word2vec word embeddings, pre-trained on the Google News dataset (300d)~\citep{mikolov2013efficient}.
\subsection{Uncertainty Estimation}
We consider two types of uncertainty as described in ~\citet{kendall2017uncertainties}: data uncertainty (aleatoric) and model uncertainty (epistemic). Data uncertainty is normally associated with properties of the data, such as imperfections in the measurements. Model uncertainty on the other hand comes from model parameters and can be explained away given enough (i.e. an infinite amount of) data. 

We also use the output of the softmax layer to measure the confidence of the model. There are four common ways to calculate uncertainty using the output of the softmax layer: Least Confidence Sampling, Margin of Confidence, Ratio of Confidence and Entropy~\citep{munro2019human}. Here we use the highest class probability as a confidence measure and refer to it as `softmax'. Using other strategies 
lead to similar conclusions (see appendices). 

\subsubsection{Data Uncertainty}
We assume aleatoric uncertainty to be a function of the data that can be learned along with the model \citep{kendall2017uncertainties}. Conceptually, this input-dependent uncertainty should be high when it is hard to predict the output given a certain input.

In order to estimate aleatoric uncertainty associated with input instances, we add an extra output to our model that represents variance $\sigma$. We then incorporate $\sigma$ into the loss function according to \citet{kendall2017uncertainties}, in the following way. 
\begin{align*}
& \sigma_i=softplus(W_\sigma u_i + b_\sigma)= ln(1+e^{W_\sigma u_i + b_\sigma})
\end{align*}
Here we assume that predictions come from a normal distribution with mean $v$ and variance $\sigma$. 
We sample $v$, distorted by Gaussian noise, $T$ times, put each through a softmax layer and pass to a standard categorical cross entropy loss function to obtain a mean over losses for all $T$ samples. 
\begin{align*}
& d_{t,i} = v_i+\sqrt{\sigma_i}*\epsilon, \epsilon \sim N(0,1)\\
& l_2 = -\frac{1}{N}\sum\limits_{n=1}^{N}\frac{1}{T}\sum\limits_{t=1}^{T}\sum\limits_{k=1}^{C}y_i^k log(softmax(d_{t,i})^k)
\end{align*}
Here $l = w_1l_1+w_2l_2$ is the total loss. 
If the original prediction $u$ was incorrect, we would need a high $\sigma$ to have varied samples away from it and hence lower the loss. In the opposite case, $\sigma$ should be small such that all samples yield a similar result, thus minimising the loss function. 
$\sigma$ is chosen as the unbound variance in logit space, which, after the model is trained, approximates input-dependent variance. 
This method can be applied to a wide range of models, but since it changes the loss function, it is likely to affect a model's performance.
\subsubsection{Model Uncertainty}
To obtain epistemic uncertainty we use the approach proposed by \citet{gal2016dropout}, which allows estimating uncertainty about a model's predictions by applying dropout at testing time and sampling from the approximate posterior. 
This approach requires no changes to the model, does not affect performance, and is relatively computationally inexpensive.
We apply dropout at testing time N times and obtain N predictions. We evaluate the differences between them to obtain a single uncertainty value in the following ways:
\paragraph{Variation Ratio}
\label{sec:epistemic}
Each of the sampled softmax predictions can be converted into an actual class label. We then define epistemic uncertainty as the proportion of cases which are not in the mode category (the label that appears most frequently). 
\begin{align*}
&v = 1-N_m/N_{total},
\end{align*}
where $N_m$ is the number of cases belonging to the mode category (most frequent class). Thus the variation ratio is 0 when all of the sampled predictions agree, indicating low model uncertainty. The upper bound would differ depending on the number of cases, but will not reach 1. 
\paragraph{Entropy}
Given an array of predictions, we average over them and then calculate predictive entropy as follows:
\begin{align*}
& s = - \sum_i p_i \log p_i.
\end{align*}
\paragraph{Variance}
Each prediction is a vector, the output of a softmax layer (entries in [0,1] which sum up to 1), of size equal to the number of classes. We calculate the variance across each dimension and then take the max value of variance as our uncertainty estimate. 
\subsection{Instance Rejection}
We assume that instances yielding high predictive uncertainty values are likely to be incorrectly predicted. We therefore make use of predictive uncertainty to filter out instances and explore the trade-off between model performance and coverage of a dataset.
We perform instance rejection in two ways; unsupervised and supervised.
\paragraph{Unsupervised}
\label{sec:unsup}
We remove portions of a dataset corresponding to instances with the highest uncertainty (separately for each uncertainty type).  
\paragraph{Supervised}
We train a supervised meta-classifier on a development set using features composed of uncertainty estimates (aleatoric, variance, entropy, variation ratio), the averaged softmax layer output and the model's prediction to decide whether an instance is correctly predicted. We reject instances classified as incorrect and evaluate performance on the rest. We compare two strong baseline models for this task: Support Vector Machines (SVM) and Random Forest (RF). 
Supervised rejection allows us to leverage all forms of uncertainty together and also dictates the number of instances to remove. 
\paragraph{Random} We have compared the two instance rejection methods above against removing portions of the test set at random. The outcome of the rejection at random does not lead to consistent performance improvement (see appendix A).
\subsection{Time-sensitive uncertainty estimates}
\begin{figure}[]
    \centering
    \includegraphics[width=\columnwidth]{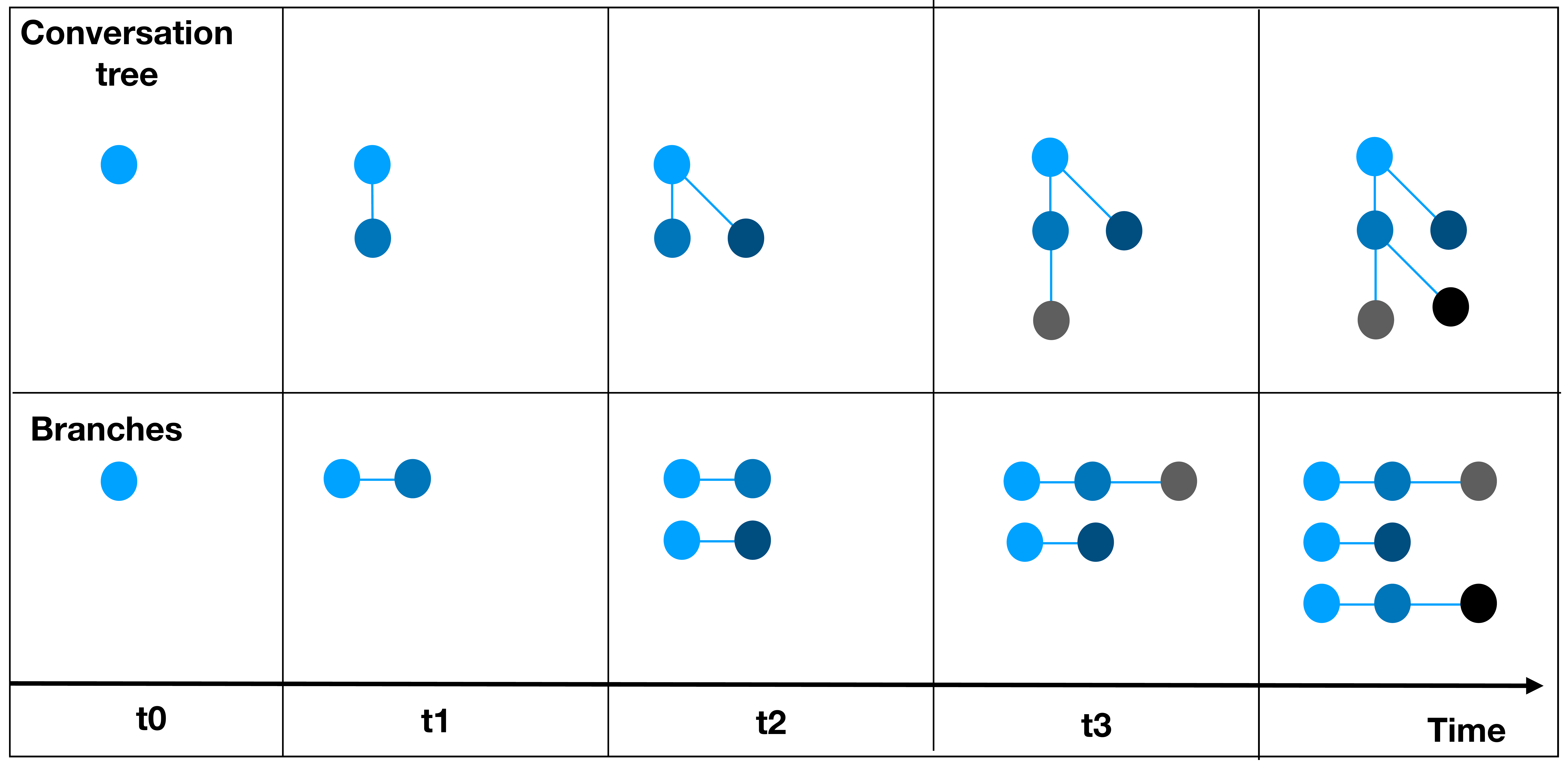}
    \caption{Development of a conversation tree over time and its decomposition into branches}
    \label{fig_time}
\end{figure}

Since rumour verification is a time-sensitive task, we have performed analysis of model uncertainty over time, as a rumour unfolds.
As illustrated in Figure~\ref{fig_time} we have deconstructed the timeline of the development of a conversation tweet by tweet, starting with just the source tweet (initiating the rumour) and adding one response at a time. 
We have then obtained model predictions and associated uncertainties for each sub-tree. As the difference between each sub-tree is a single tweet, we can track the development of uncertainty alongside the development of a conversation, and the effect each added response has.
\subsection{Calibration}
Uncertainty estimates obtained do not correspond to the actual probabilities of the prediction being correct, they instead order the samples from the least likely to be correct to the most likely. While the order provided by the scores is sufficient for unsupervised and supervised rejection, these scores can be on a different scale for different datasets and do not allow for direct comparison between models, i.e. they are not calibrated.
Calibration refers to a process of adjusting confidence scores to correspond to class membership probabilities, i.e if $N$ predictions have a confidence of 0.5, then 50\% of them should be correctly classified in a perfectly calibrated case. 
Modern neural networks are generally poorly calibrated 
and hyper-parameters of the model 
 influence the calibration~\citep{guo2017calibration}. MC dropout uncertainty is thus also influenced by hyperparameters but can be calibrated using dropout probability~\cite{gal2016uncertainty}.

To evaluate how well confidence scores are calibrated, one can use reliability diagrams and Expected Calibration Error (ECE) scores ~\citep{guo2017calibration}. ECE is obtained by binning $n$ confidence scores into $M$ intervals and comparing the accuracy of each bin against the expected one in a perfectly calibrated case (equal to the confidence of the bin): $ECE = \sum_{m=1}^{M}\frac{|B_m|}{n}|acc(B_m)-conf(B_m)|$.
Confidence calibration can be improved using Calibration methods. These are post-processing steps that produce a mapping from existing scores to calibrated probabilities using a held-out set. Common approaches are Histogram binning, Isotonic regression and Temperature scaling~\citep{guo2017calibration}.
\begin{table}[]
\resizebox{\columnwidth}{!}{%
\begin{tabular}{|l|l|l|l|l|l|l|}
\hline
 &\# Posts &  \# Trees  & T & F & U & NR \\ \hline
PHEME  & 33288 & 2410 & 1067 & 639 & 704 & 0 \\ \hline
Twitter 15  & 40927 & 1374 & 350 & 336 & 326 & 362 \\ \hline
Twitter 16  & 18770 & 735 & 189 & 173 & 174 & 199 \\ \hline
\end{tabular}%
}
\caption{Number of posts, conversation trees and class distribution in the datasets (T -- True, F -- False, U -- Unverified, NR -- Non-Rumour).}
\label{tab:datastats}
\end{table}
\section{Data}
In our experiments we use publicly available datasets of Twitter conversations discussing rumours. Table \ref{tab:datastats} shows the number of conversation trees in the datasets and the class distribution.
\subsection{PHEME}
We use conversations from the PHEME dataset discussing rumours related to nine newsbreaking events. Rumours in this dataset were labeled as True, False or Unverified by professional journalists ~\citep{zubiaga2016analysing}. When conducting experiments on this dataset we perform cross-validation in a leave-one-event-out setting, i.e. using all the events except for one as training, and the remaining event as testing. This is a challenging setup, imitating a real-world scenario, where a model needs to generalise to unseen rumours. The number of rumours, the number of the corresponding conversations, as well as the class label distribution (true-false-unverified) vary greatly across events.
\subsection{Twitter 15/16}
The Twitter 15 and Twitter 16 datasets were made publicly available by \citet{ma2017detect}, and were created using reference datasets from Ma\citet{ma2016detecting} and \citet{liu2015real}. Claims were annotated using veracity labels on the basis of articles corresponding to the claims found in rumour debunking websites such as \textit{snopes.com} and  \textit{emergent.info}. These datasets merge rumour detection and verification into a single four-way classification task, containing True, False and Unverified rumours as well as Non-Rumours. Both datasets are split into 5 folds for cross validation, and contrary to the PHEME dataset, folds are of approximately equal size with a balanced class distribution.
\begin{figure*}[h]
   \subfloat[PHEME]{%
      \includegraphics[width=0.32\textwidth]{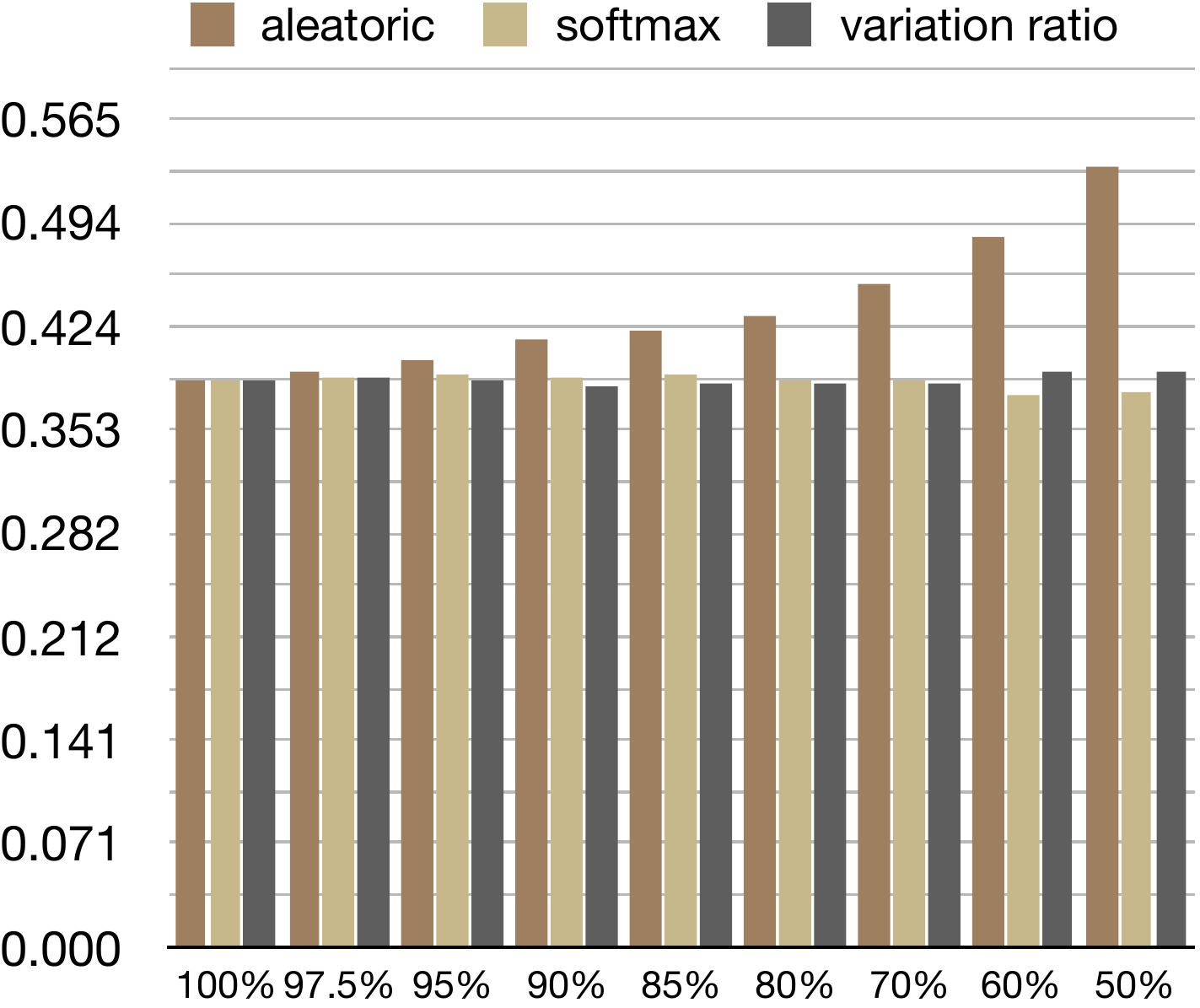}}
\hspace{\fill}
   \subfloat[Twitter 15]{%
      \includegraphics[width=0.32\textwidth]{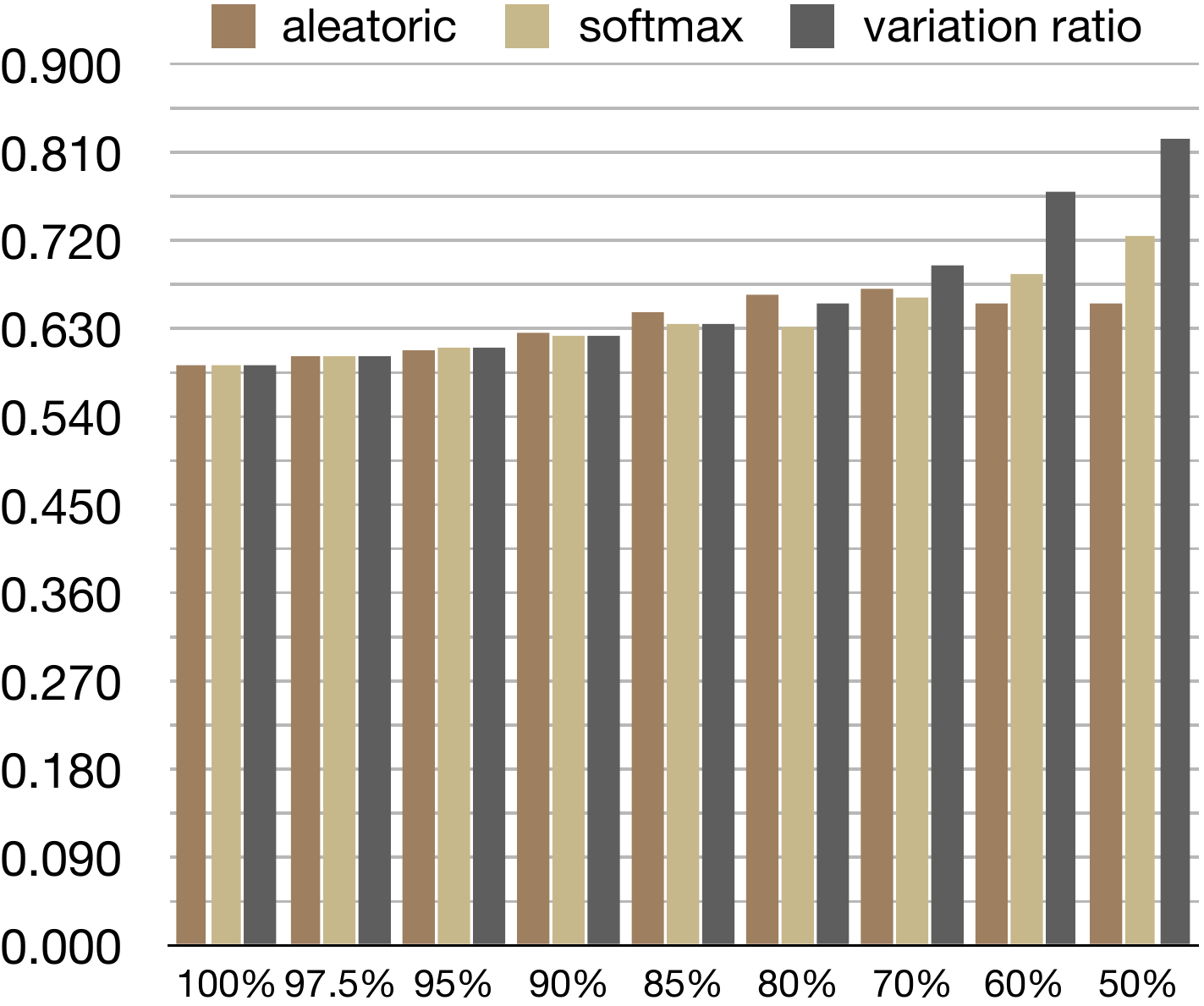}}
\hspace{\fill}
   \subfloat[Twitter 16]{%
      \includegraphics[width=0.32\textwidth]{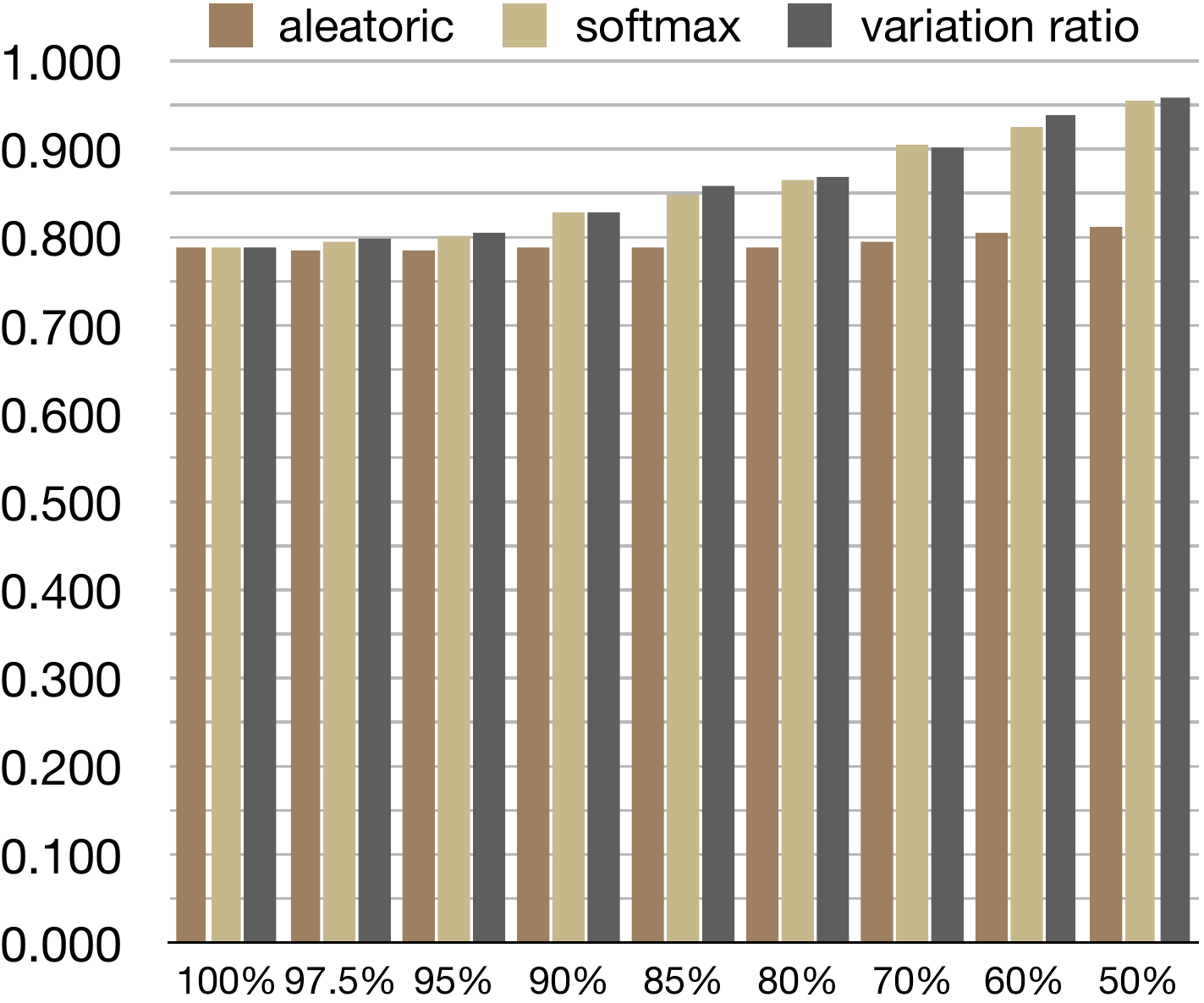}}\\
\caption{Unsupervised rejection of instances with the highest uncertainty and corresponding lowest confidence (softmax) values across 3 datasets. The Y-axis shows performance in terms of accuracy, on the X-axis the percentage of the remaining instances is shown.}
\label{fig_unsup_rej}
\end{figure*}
\begin{table*}[h]
\centering
\resizebox{\textwidth}{!}{%
\begin{tabular}{|l|l|l|l|l|l|l|l|l|l|l|l|l|}
\hline
\multirow{3}{*}{} & \multicolumn{2}{c|}{\multirow{2}{*}{All instances}} & \multicolumn{1}{c|}{\multirow{2}{*}{Classifier}} & \multicolumn{1}{c|}{\multirow{2}{*}{N removed}} & \multicolumn{2}{c|}{\multirow{2}{*}{Supervised rejection}} & \multicolumn{6}{c|}{Unsupervised rejection} \\ \cline{8-13} 
 & \multicolumn{2}{c|}{} & \multicolumn{1}{c|}{} & \multicolumn{1}{c|}{} & \multicolumn{2}{c|}{} & \multicolumn{2}{c|}{aleatoric} & \multicolumn{2}{c|}{epistemic (variation ratio)} & \multicolumn{2}{c|}{softmax} \\ \cline{2-13} 
 & \multicolumn{1}{c|}{Accuracy} & \multicolumn{1}{c|}{Macro F} & \multicolumn{1}{c|}{} & \multicolumn{1}{c|}{} & \multicolumn{1}{c|}{Accuracy} & \multicolumn{1}{c|}{Macro F} & \multicolumn{1}{c|}{Accuracy} & \multicolumn{1}{c|}{Macro F} & \multicolumn{1}{c|}{Accuracy} & \multicolumn{1}{c|}{Macro F} & \multicolumn{1}{c|}{Accuracy} & \multicolumn{1}{c|}{Macro F} \\ \hline
\multirow{2}{*}{PHEME} & \multirow{2}{*}{0.278} & \multirow{2}{*}{0.225} & SVM & 1057 & \textbf{0.399} & 0.196 & 0.306 & 0.216 & 0.35 & 0.235 & 0.332 & 0.239 \\ \cline{4-13} 
 &  &  & RF & 1179 & \textbf{0.378} & 0.235 & 0.311 & 0.217 & 0.346 & 0.227 & 0.329 & 0.236 \\ \hline
\multirow{2}{*}{Twitter 15} & \multirow{2}{*}{0.671} & \multirow{2}{*}{0.67} & SVM & 402 & \textbf{0.806} & \textbf{0.801} & 0.656 & 0.632 & 0.801 & 0.795 & 0.794 & 0.788 \\ \cline{4-13} 
 &  &  & RF & 504 & \textbf{0.834} & \textbf{0.829} & 0.662 & 0.624 & 0.836 & 0.828 & 0.818 & 0.811 \\ \hline
\multirow{2}{*}{Twitter 16} & \multirow{2}{*}{0.755} & \multirow{2}{*}{0.756} & SVM & 184 & \textbf{0.895} & \textbf{0.893} & 0.751 & 0.744 & 0.885 & 0.878 & 0.878 & 0.868 \\ \cline{4-13} 
 &  &  & RF & 197 & \textbf{0.897} & \textbf{0.892} & 0.755 & 0.747 & 0.887 & 0.878 & 0.884 & 0.873 \\ \hline
\end{tabular}%
}
\caption{How rejecting instances using supervised and unsupervised methods affects model performance across datasets, in terms of both accuracy and macro F-score.  Performance values were obtained in a separate set of experiments, by removing one of the folds from the training set, as supervised models needed an extra development set to be trained on.}
\label{tab_sup_rej}
\end{table*}
\section{Experimental Setup}
\label{sec_setup}
We perform cross-validation on all of the datasets. When choosing parameters, we choose one of the folds within each dataset to become the development set: CharlieHebdo in PHEME (large fold with balanced labels) and fold 0 in Twitter 15 and Twitter 16. We evaluate models using both accuracy and macro F-score due to the class imbalance in the PHEME dataset\footnote{\url{https://github.com/kochkinaelena/Uncertainty4VerificationModels}}. During the cross-validation iterations each fold becomes a testing set once. We then aggregate model predictions from each fold, resulting in predictions for the full dataset, and use them to perform evaluation as well as unsupervised instance rejection based on uncertainty levels.

To perform supervised rejection we need to train a meta-classifier on a subset of data that was not used for training the rumour verification model. Therefore in a separate set of experiments we exclude one of the folds (development set) from training of the verification model. We run cross-validation with one less fold and at each step obtain predictions and uncertainty estimates for both the test fold and the development set. We then use the predictions and uncertainty values predicted for the instances in the development set as training instances in our rejection meta-models, which we then evaluate on each of the corresponding test folds, thus obtaining the combined predictions for all of the folds in the dataset except for the development. This set up corresponds to results shown in Table \ref{tab_sup_rej}, as one of the folds was removed from training. The results are therefore not directly comparable to the ones in Figure \ref{fig_unsup_rej} or in previous literature \citep{kochkina2018all,ma2018rumor}.
\begin{figure*}[]
   \subfloat[]{%
      \includegraphics[width=0.3\textwidth]{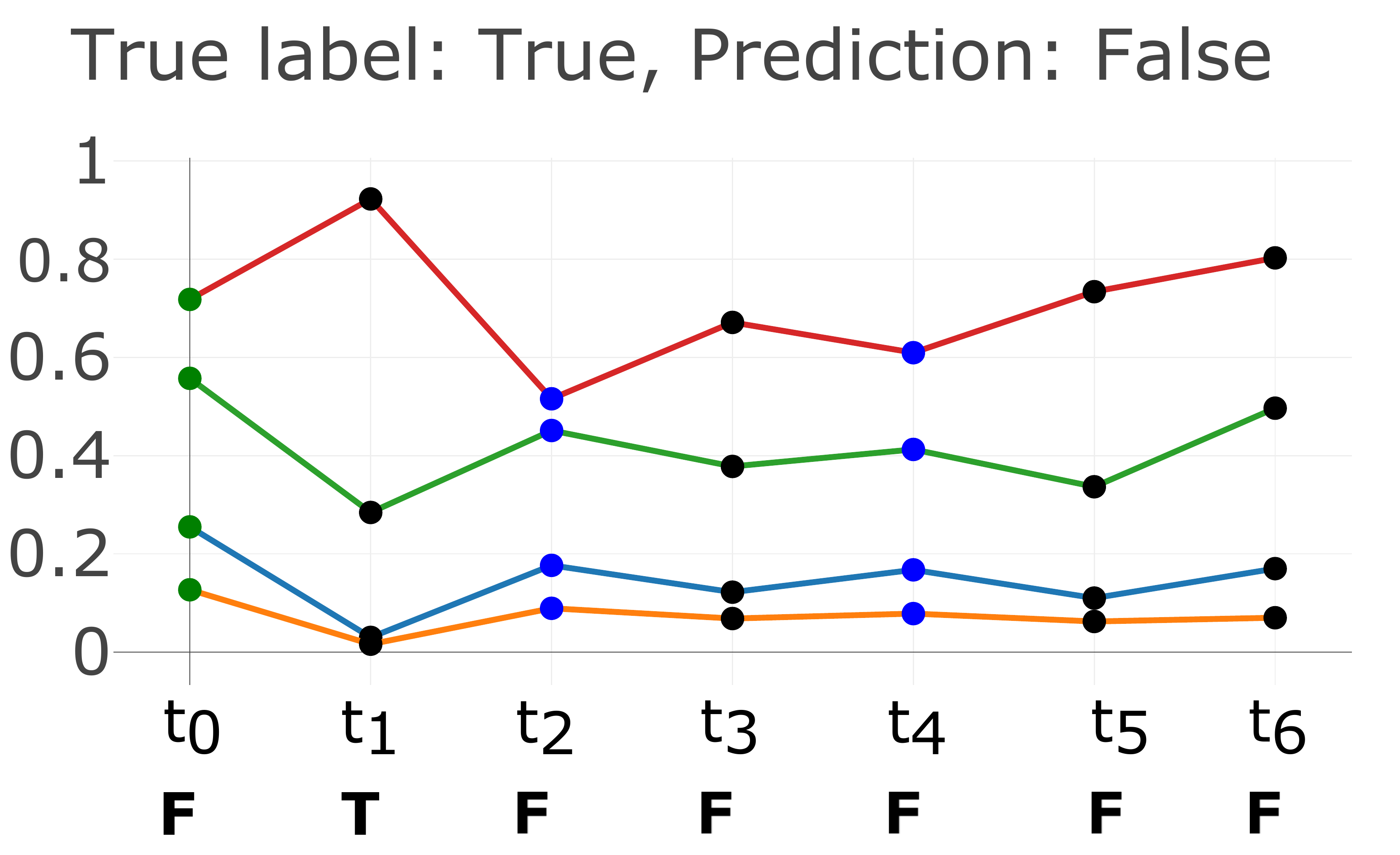}}
\hspace{\fill}
   \subfloat[]{%
      \includegraphics[width=0.3\textwidth]{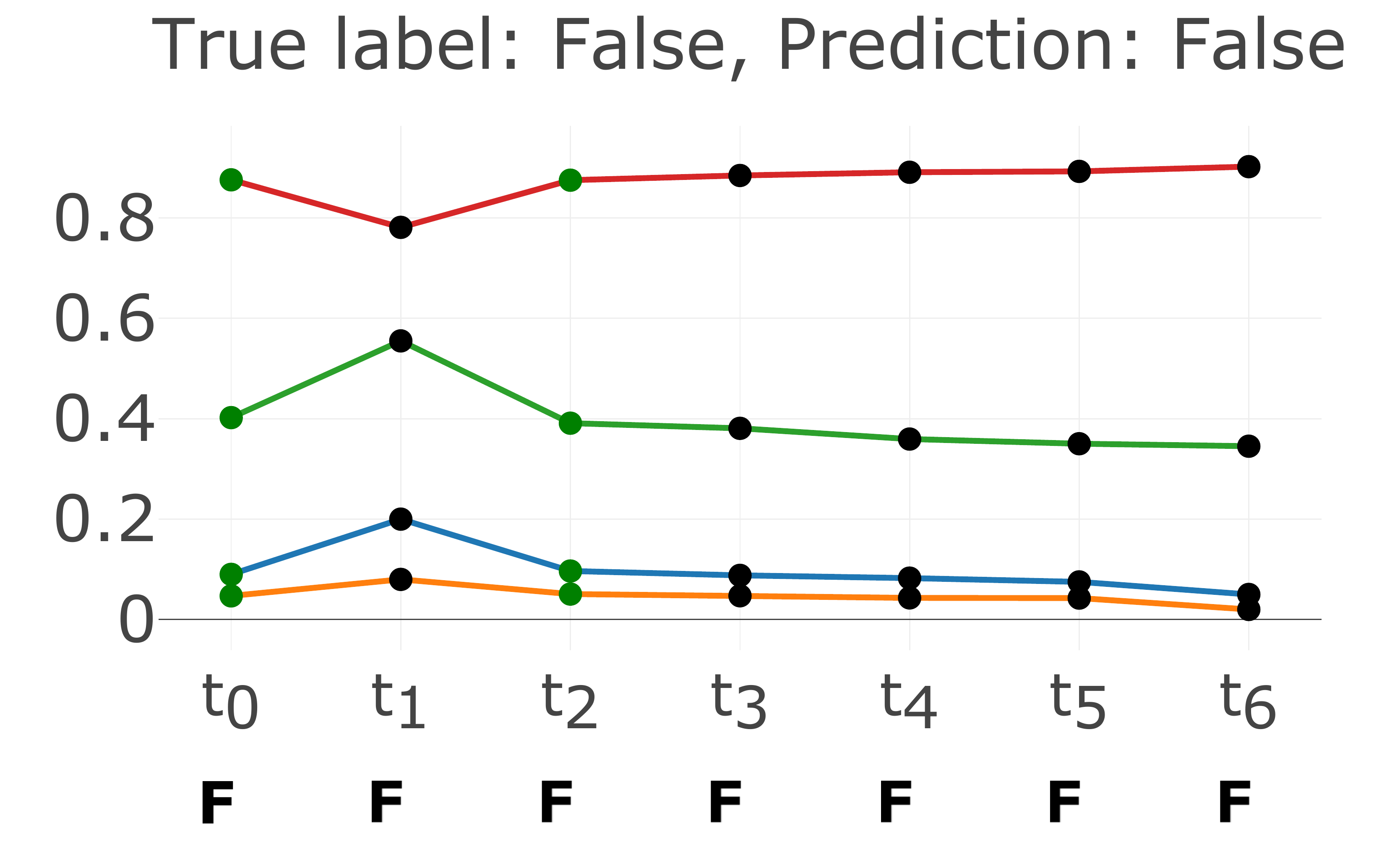}}
\hspace{\fill}
   \subfloat[]{%
      \includegraphics[width=0.37\textwidth]{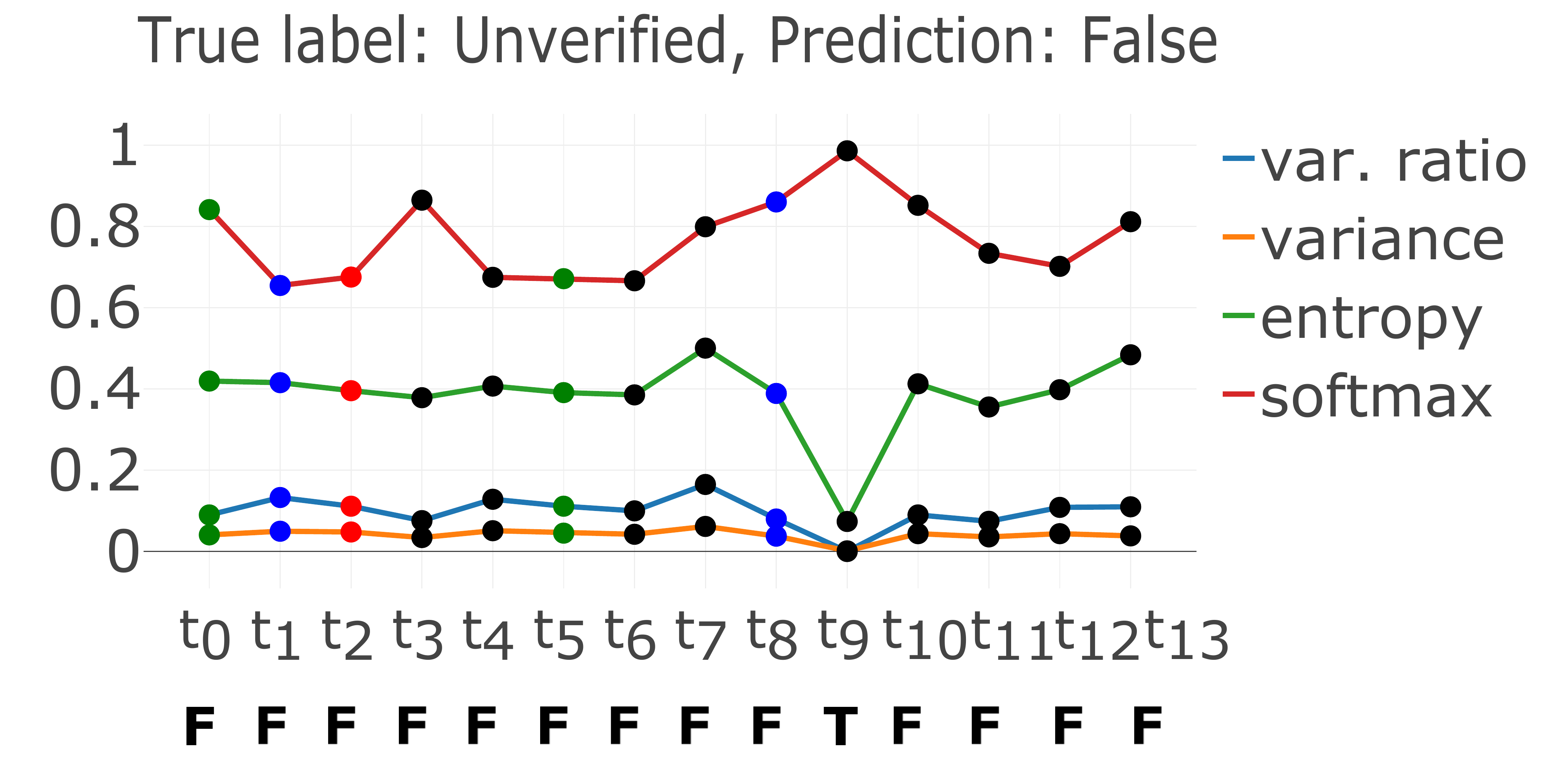}}\\
 \subfloat[]{%
      \includegraphics[width=0.3\textwidth]{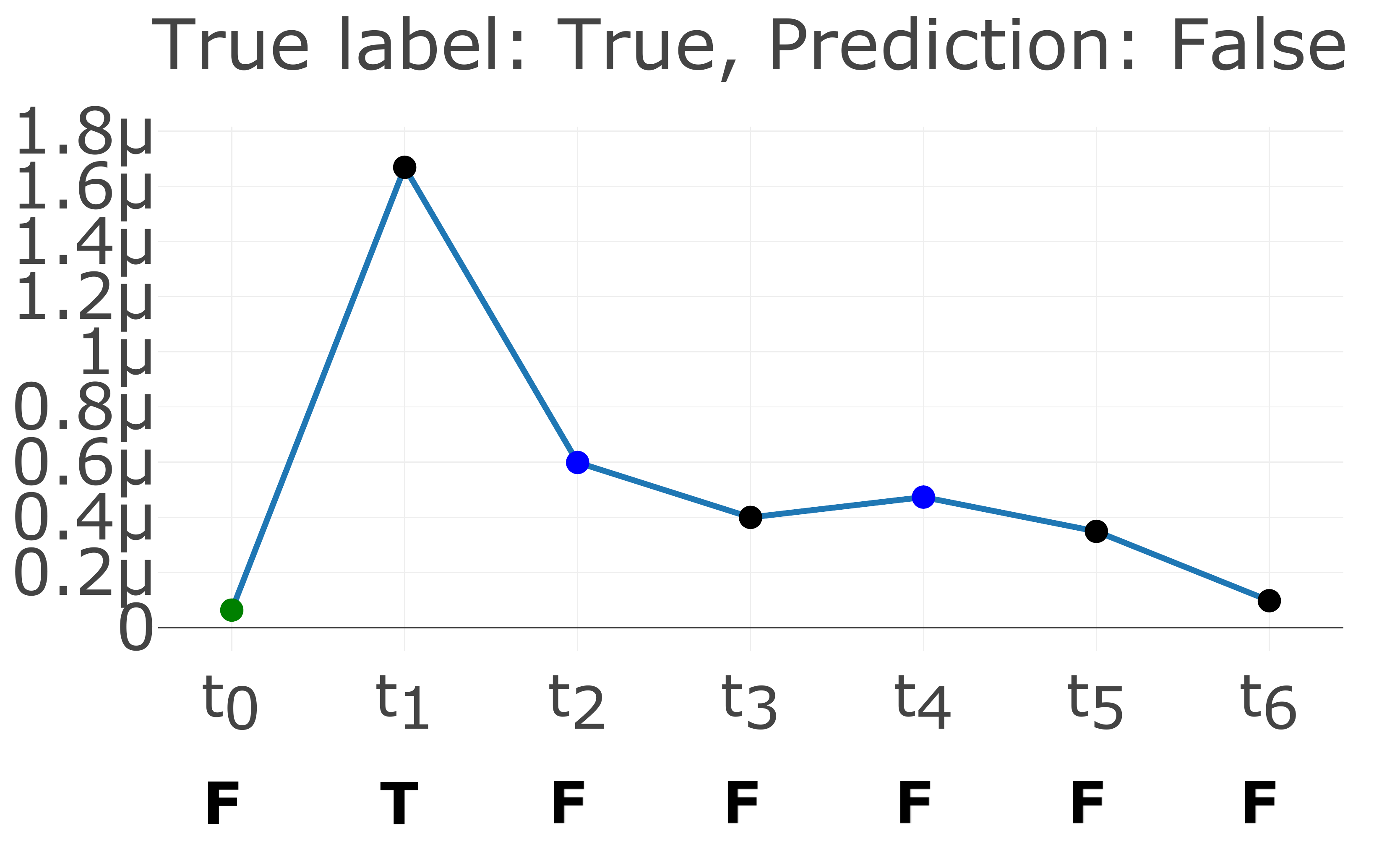}}
\hspace{\fill}
   \subfloat[]{%
      \includegraphics[width=0.3\textwidth]{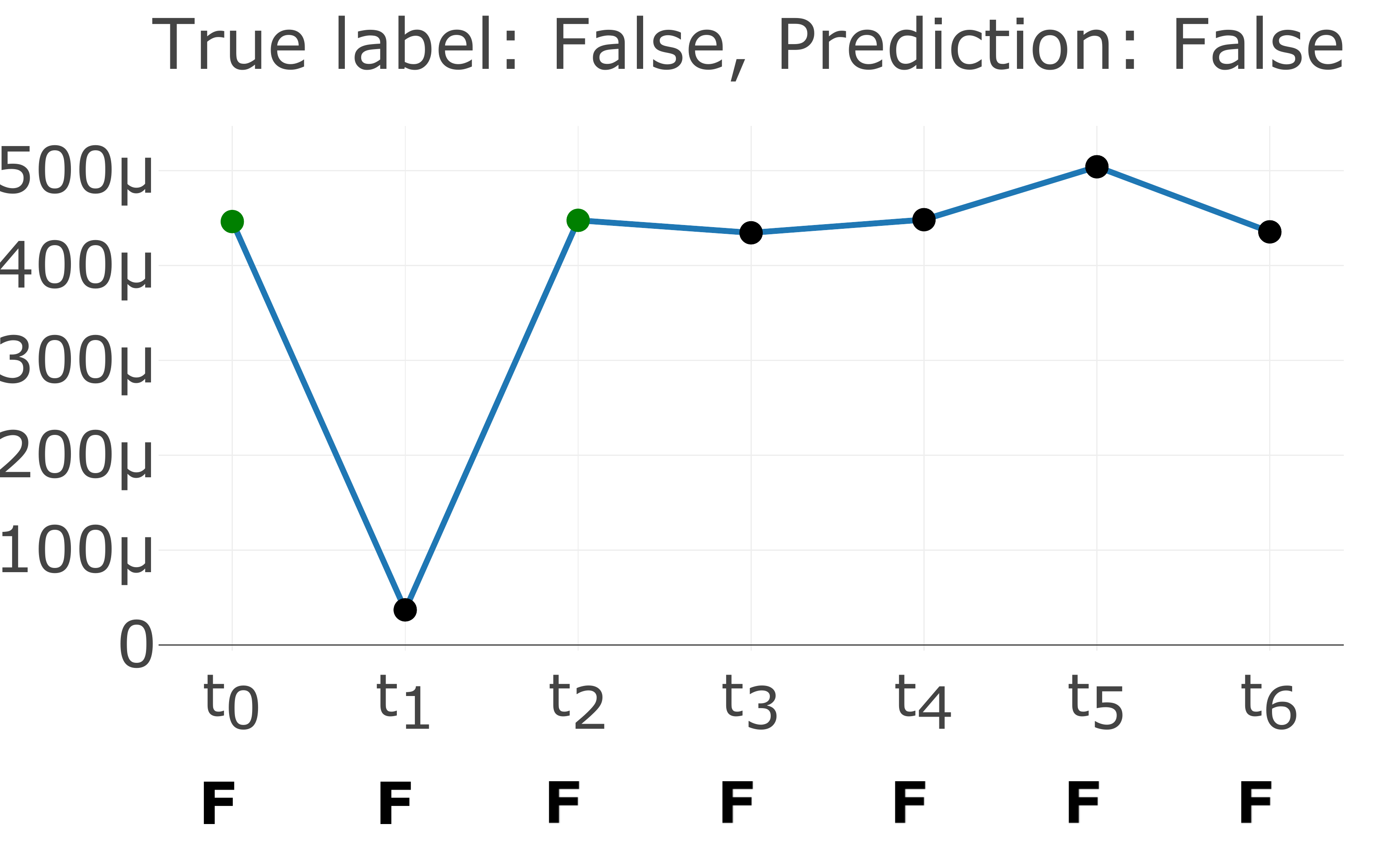}}
\hspace{\fill}
   \subfloat[]{%
      \includegraphics[width=0.37\textwidth]{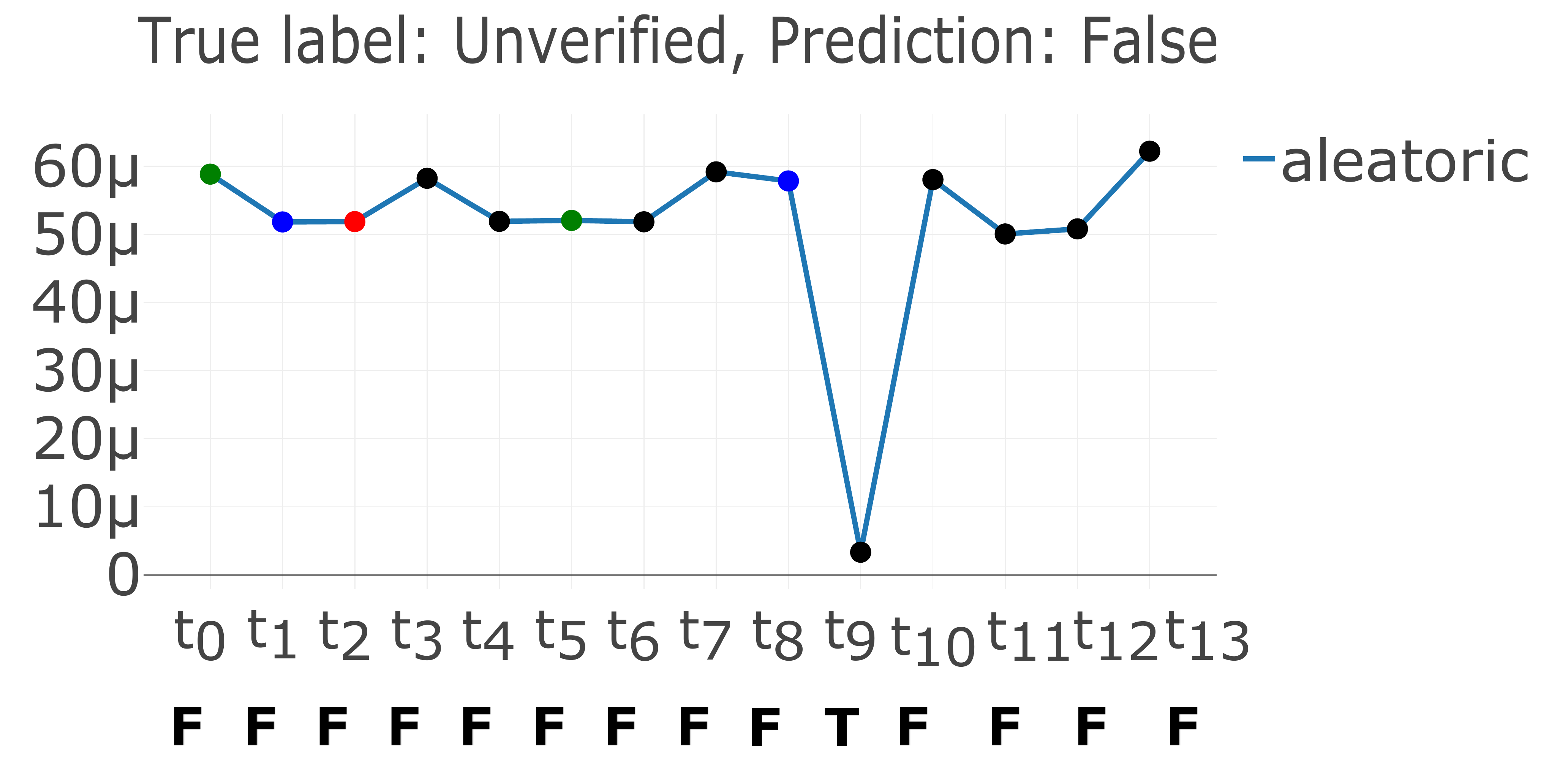}}
\caption{Examples of uncertainty development over time for three conversations discussing rumours from the PHEME dataset. Each of the nodes is labeled with its predicted stance label: green -- supporting, red -- denying, blue -- questioning and black -- commenting. Predictions are in bold at the bottom, where F -- False, T -- True, U -- Unverified.}
\label{fig_uncerttime}
\end{figure*}
\section{Results}
\label{results_section}
\subsection{Unsupervised Rejection}
Figure \ref{fig_unsup_rej} shows the effect of applying unsupervised rejection (as explained in section \ref{sec:unsup}). Each plot shows model performance in terms of accuracy, where the first bar of each plot shows model performance with all instances present and the following bars show performance for the corresponding percentage of remaining instances. 
Figure \ref{fig_unsup_rej} shows the effect of unsupervised rejection using aleatoric and epistemic uncertainty (calculated as variation ratio, see section ~\ref{sec:epistemic})\footnote{We performed experiments using variance and entropy values with similar outcomes (appendix A).}, as well as the softmax class probabilities as a measure of confidence (1-uncertainty). 
Initial performance using $100\%$ of the data (Figure~\ref{fig_unsup_rej}) on the PHEME dataset is markedly different to Twitter 15,16 due to the dataset and task-setup differences. On the Twitter 15 dataset branch-LSTM does not reach the state-of-the-art Tree-GRU ~\citep{ma2018rumor}, however branch-LSTM outperforms Tree-GRU on the Twitter 16 dataset. On the PHEME dataset performance is comparable and slightly improved over the results in \citet{kochkina2018all}.
In line with model performance, the effect of rejection using aleatoric and epistemic uncertainties is different for PHEME compared to Twitter 15,16.  Figure~\ref{fig_unsup_rej} (a) shows that in PHEME greater improvement in accuracy comes from using aleatoric uncertainty, whereas for Twitter 15 (b) and Twitter 16 (c) there is very little improvement with aleatoric uncertainty compared to epistemic. We believe this is due to the nature of the datasets: folds in PHEME differ widely in size and class balance, resulting in higher/more varied data uncertainty values, in contrast with the very balanced datasets of Twitter 15,16.  The effect of rejection using low values of softmax confidence is also positive and often similar to the effect of epistemic uncertainty as it is also estimating model's uncertainty. However softmax is outperformed by other types of uncertainty in most cases (Figure~\ref{fig_unsup_rej}). 
\subsection{Supervised Rejection}
Table~\ref{tab_sup_rej} shows the comparison of two models for supervised rejection versus unsupervised rejection of the same number of instances for all three datasets. Note that performance value in Table~\ref{tab_sup_rej} differs from that in Figure~\ref{fig_unsup_rej} as this was obtained in a separate set of experiments (as described in section~\ref{sec_setup}). Having less training data harmed performance on PHEME and Twitter 16. 
Table~\ref{tab_sup_rej} shows that using supervised rejection is better than unsupervised in terms of accuracy scores for all datasets and also in terms of macro F-scores for the Twitter 15,16 datasets. We believe that the reason the same effect on macro-F score is not observed in PHEME is the class imbalance in this dataset.

Comparing the two methods, SVM and RF, for supervised rejection we observe that RF leads to a larger amount of instances being removed, achieving higher performance than SVM. However, the difference in performance between the two is very small. As part of future work the meta-classifier can be improved further, made more complex or incorporated in the predictive model, making it closer to active learning, closing the loop from prediction and corresponding uncertainty to classifier improvement.
Another benefit of using a supervised model for instance  rejection is that it can be further tuned, e.g., by varying the threshold boundary to prioritise high precision over recall. The precision value of this meta-classifier is the same as the accuracy of the predictions obtained after the rejection procedure. 
\begin{figure*}[h]
 \subfloat[Epistemic PHEME]{%
      \includegraphics[width=0.32\textwidth]{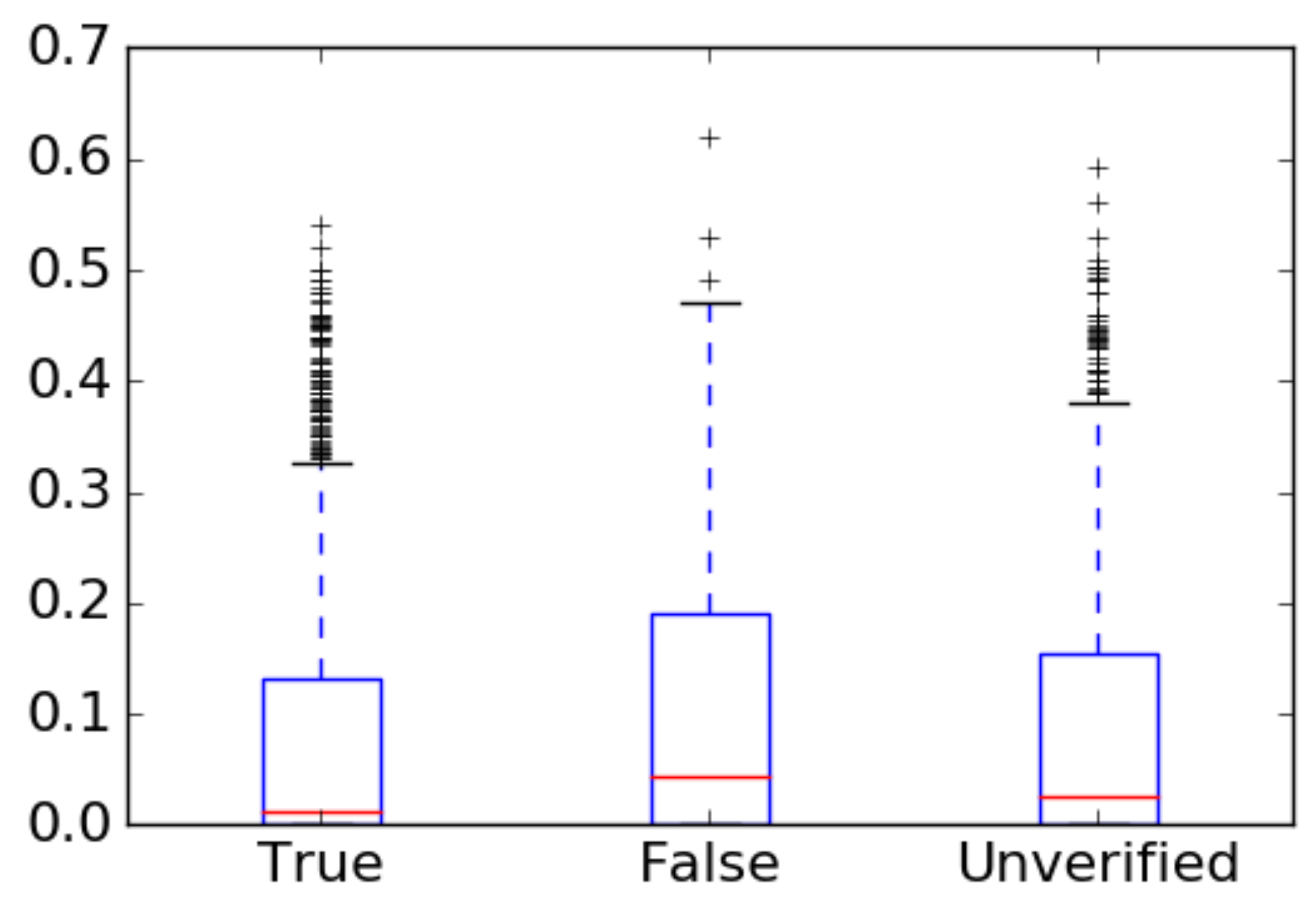}}
\hspace{\fill}
   \subfloat[Epistemic Twitter 15]{%
      \includegraphics[width=0.32\textwidth]{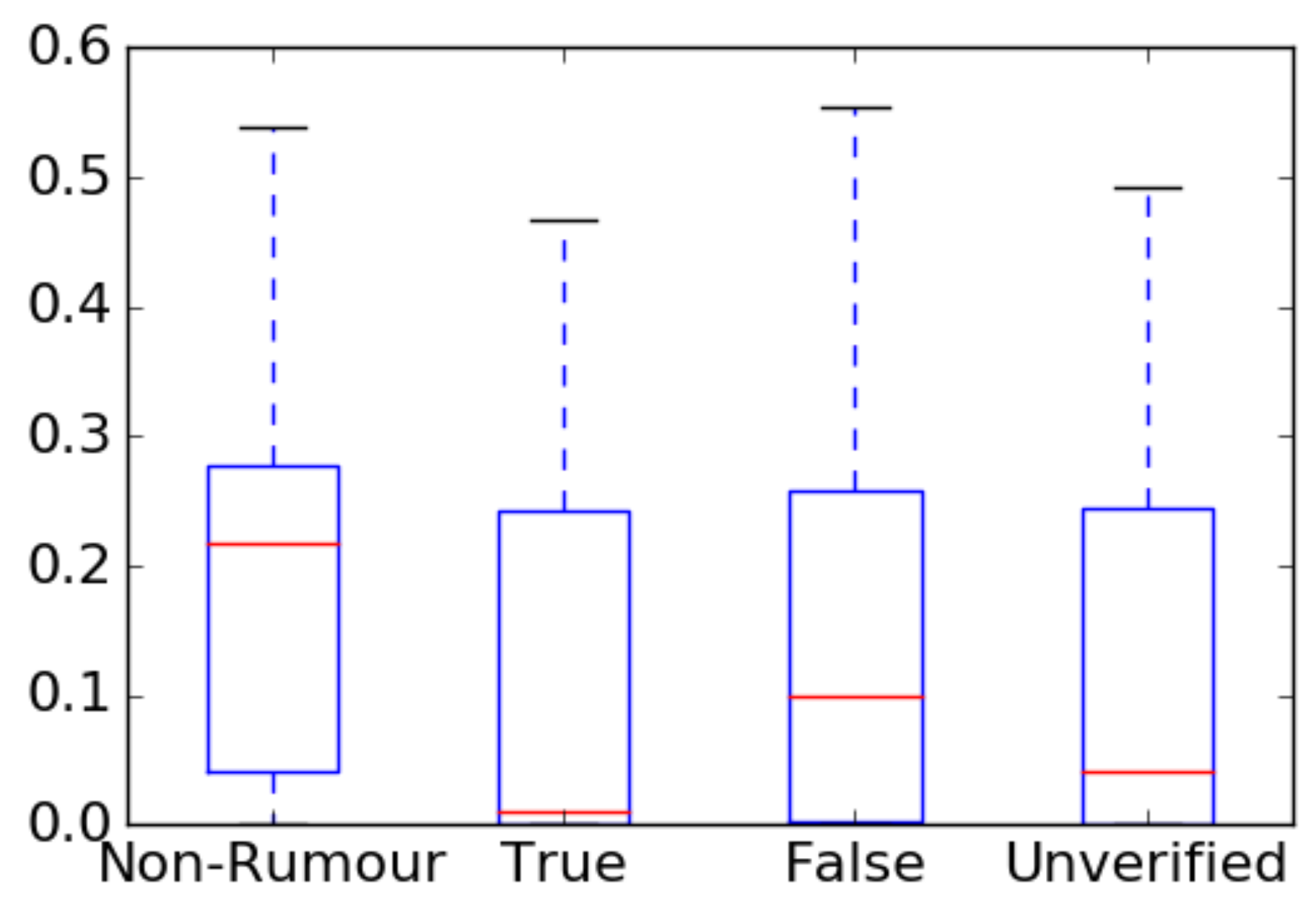}}
\hspace{\fill}
   \subfloat[Epistemic Twitter 16]{%
      \includegraphics[width=0.32\textwidth]{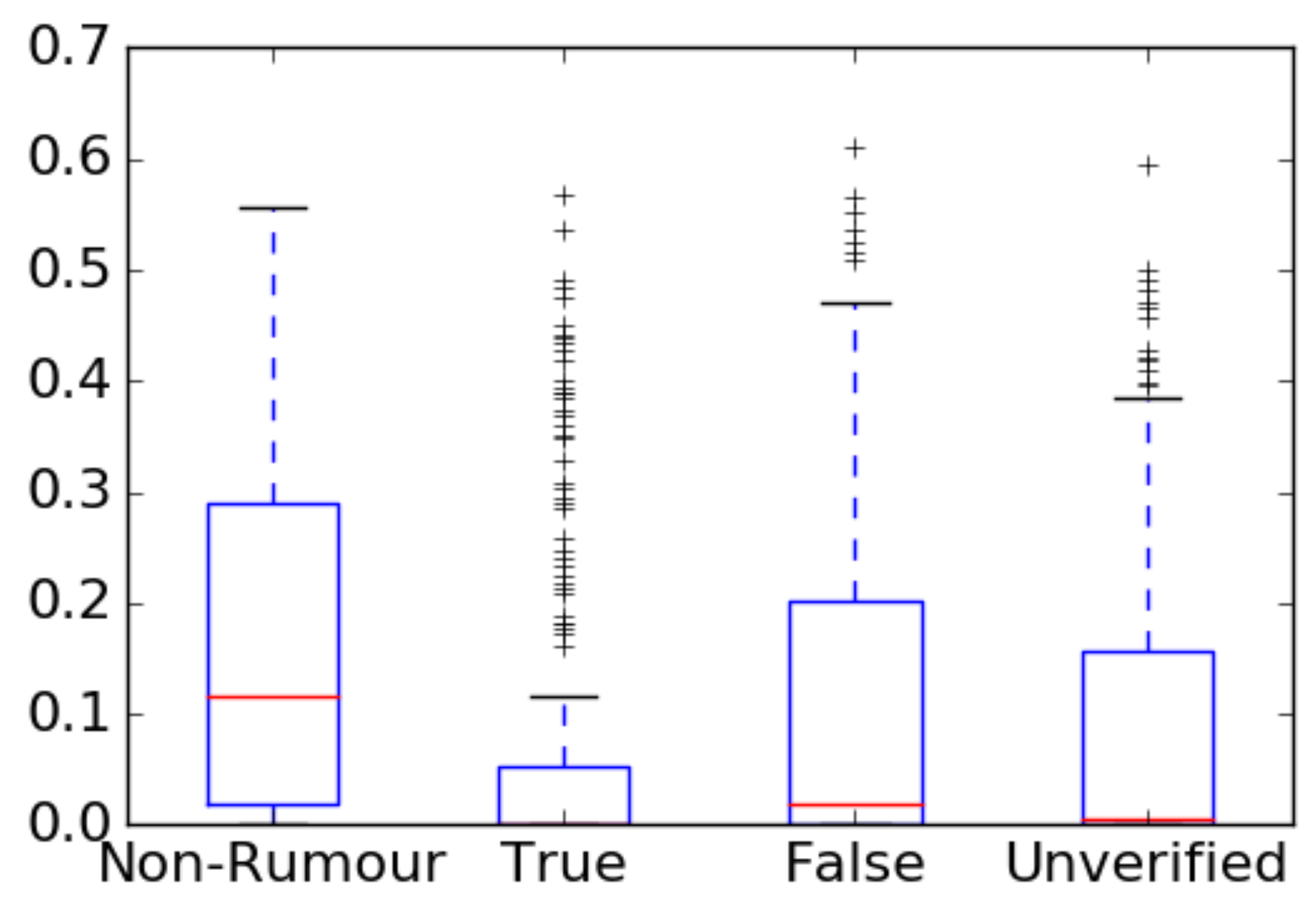}}\\     
\caption{Effect of class labels on uncertainty estimates.}
\label{fig_class_box}
\end{figure*}
\begin{table}[]
\centering
\resizebox{\columnwidth}{!}{%
\begin{tabular}{|l|l|l|l|l|}
\hline
           & True  & False & Unverified & Non-Rumour \\ \hline
PHEME      & 0.569 & 0.198 & 0.163      & -          \\ \hline
Twitter 15 & 0.679 & 0.618 & 0.608      & 0.503      \\ \hline
Twitter 16 & 0.88  & 0.729 & 0.755      & 0.739      \\ \hline
\end{tabular}%
}
\caption{Per-class f1-scores of branch-LSTM model on each of the datasets.}
\label{tab:perclass}
\end{table}
\subsection{Timeline analysis}
Part of the PHEME dataset was annotated for stance ~\citep{derczynski2017semeval}. We used the open-source branch-LSTM model trained on that part to obtain predicted stance labels for the rest of the PHEME dataset~\citep{kochkina2017turing}. 
There is no stance information for the Twitter 15,16 datasets, so this analysis is only available for the PHEME dataset. Note that we did not provide stance as a feature to train the veracity classifier: we assume that stance is an implicit feature within the tweets. 
Figure~\ref{fig_uncerttime} shows examples of timelines of changes in predictions and uncertainty levels over time.  Sub-plots (a) -- (c) show all types of epistemic uncertainty: variation ratio (blue), entropy (green), variance (orange) as well as softmax confidence (red); on sub-plots (d) -- (f) we show aleatoric uncertainty of the conversations corresponding to the above plots separately, as values are on a different scale. Each of the nodes is labeled with its predicted stance label: green -- supporting, red -- denying, blue -- questioning and black -- commenting. 
One could expect to see uncertainty decreasing over time as more information about a rumour becomes available (we can see this effect only very weakly on sub-plot Figure~\ref{fig_uncerttime}(b), showing a correctly predicted False rumour). However, not all responses are equally relevant and also the stance of new posts varies, therefore the uncertainty levels also change. 
Interestingly, the true rumour on subplot Figure~\ref{fig_uncerttime}(a) (incorrectly predicted as False during the final time steps) had low uncertainty at step 2 and was predicting a correct label. However, the model appears to have been confused by further discussion resulting in an incorrect prediction with higher uncertainty levels. 
The analysis of uncertainty as a rumour unfolds can be used not only to analyse the effect of stance but also to study other properties of rumour spread.
Only $5-20\%$ of the conversations have a change in predictions as the conversation unfolds suggesting that source tweets are the most important for the model. Furthermore, we can use the timelines of uncertainty measurements in order to only allow predictions at the time steps with lowest uncertainty, which may lead to performance improvements. In experiments with the PHEME dataset accuracy grew from 0.385 to 0.395 using variation ratio and to 0.398  using aleatoric uncertainty estimates. 

When analysing the relation between uncertainty and the conversation size, we observed that for the confidence levels represented by the output of the softmax layer, conversations with a larger amount of tweets had higher uncertainty. However, for aleatoric and epistemic estimates we do not observe a strong trend of uncertainty increase with the size of the conversation (see box plots in appendix D), which would indicate that these types of uncertainty are more robust in this respect. 
Higher levels of uncertainty associated with longer conversations may be due to the fact that responses became less informative and/or conversation changed topic. They may also be stemming from a weakness in model architecture in terms of its ability to process long sequences. 
\subsection{Uncertainty and Class Labels}
Is higher uncertainty associated with a particular class label?
Figure~\ref{fig_class_box} shows boxplots of epistemic uncertainty values associated with each of the three classes in the PHEME dataset and each of the four classes in Twitter 15,16. Table~\ref{tab:perclass} shows per-class model performance on the full datasets.
In all datasets the True class has significantly lower levels of uncertainty (using \citet{kruskal1952use} test between the groups), while the uncertainties for False and Unverified are higher than True. The difference between False and Unverified is not statistically significant in any cases.
Aleatoric uncertainty shows a similar pattern for the class labels. In Twitter 15,16 the Non-Rumour class has the highest uncertainty (and relatively lower f1 score). 
These outcomes are inline with findings in \citet{kendall2019geometry} which showed an inverse relationship between uncertainty and class accuracy or class frequency.
\subsection{Calibration outcomes}
We measure and compare the ECE for all types of uncertainty. We apply Histogram Binning, a simple yet effective approach to improve the calibration for each type of uncertainty. We use the experiment setup with one of the folds reserved as development set to train the calibration method. We convert uncertainty estimates $u$ into confidence scores as $1-u$, and for aleatoric uncertainty we normalise it to be in $[0,1]$. 
Table \ref{tab:ece} shows the ECE before and after calibration, for different uncertainty measures -Softmax (S), Aleatoric (A), Variation Ratio (VR)- where a lower value indicates better calibration (calibration curves can be found in appendix E). Initial ECE for PHEME is higher than for Twitter 15 and 16 datasets. 
VR has the best initial calibration, however Histogram Binning notably improves calibration across all datasets and uncertainty types. 

\begin{table}[]
\centering
\resizebox{\columnwidth}{!}{%
\begin{tabular}{|l|l|l|l|l|l|l|}
\hline
\multirow{2}{*}{} & \multicolumn{3}{l|}{No calibration} & \multicolumn{3}{l|}{Histogram Binning} \\ \cline{2-7} 
                  & S          & A          & VR        & S           & A           & VR         \\ \hline
PHEME             & 0.646      & 0.683      & 0.492     & 0.173       & 0.088       & 0.111      \\ \hline
Twitter 15        & 0.265      & 0.333      & 0.216     & 0.056       & 0.039       & 0.062      \\ \hline
Twitter 16        & 0.191      & 0.196      & 0.121     & 0.164       & 0.079       & 0.044      \\ \hline
\end{tabular}%
}
\caption{Expected Calibration Error before and after applying calibration over uncertainty estimates. S -- softmax (LCS), A -- aleatoric uncertainty, VR -- variation ratio. }
\label{tab:ece}
\end{table}
\section{Discussion}
We have shown that data and model uncertainties can be included as part of the evaluation of any deep learning model without harming its performance. Moreover, even though data uncertainty estimation changes the loss function of a model, it often leads to improvements \citep{kendall2017uncertainties}. 
When performing rejection in an unsupervised fashion we need to know when to stop removing instances. Defining a threshold of uncertainty is not straightforward as uncertainty will be on a different scale for different datasets. Supervised rejection leverages all forms of uncertainty together and dictates the number of instances to remove. Thus to tune both methods availability of a development set is important. 

 While we are not focusing on user uncertainty here, in rumour verification linguistic markers of user uncertainty (words like ``may", ``suggest", ``possible") are  associated with rumours. In the PHEME dataset such expressions often occur in unverified rumours, thus conversations containing them are easier to classify, and hence they are associated with lower predictive uncertainty.

\section{Conclusions and Future Work}
We have presented a method for obtaining model and data uncertainty estimates on the task of rumour verification in Twitter conversations. We have demonstrated two ways in which uncertainty estimates can be leveraged to remove instances that are likely to be incorrectly predicted, so that making a decision concerning those instances can be prioritised by a human. We have also shown how uncertainty estimates can be used to interpret model decisions over time.
Our results indicate that the effect of data uncertainty and model uncertainty varies across datasets due to differences in their respective properties. The methods presented here can be selected based on knowledge of the properties of the data at hand, for example prioritising the use of aleatoric uncertainty estimates on imbalanced and heterogeneous datasets such as PHEME. For best results, one should use a combination of aleatoric and epistemic uncertainty estimates and tune the parameters of uncertainty estimation methods using a development set.
Using uncertainty estimation methods can help identify which instances are hard for the model to classify, thus highlighting the areas where one should focus during model development. 

Future work would include a comparison with other, more complex, methods for uncertainty estimation, incorporating uncertainty to affect model decisions over time, and further investigating links between uncertainty values and linguistic features of the input.

\section*{Acknowledgements}
This work was supported by The Alan Turing Institute under the EPSRC grant EP/N510129/1. 
\bibliography{acl2020}
\bibliographystyle{acl_natbib}

\appendix
\section{Comparison of unsupervised rejection performance using each type of uncertainty versus random rejection}

Tables \ref{unsup}-\ref{unsuptw16} present the results in terms of accuracy of unsupervised rejection of instances with the highest uncertainty and corresponding lowest confidence (softmax) values against random rejection of instances across 3 datasets: PHEME, Twitter 15, Twitter 16.

In all cases random rejection does not lead to consistent performance improvements, and hence, is outperformed by (un)certainty-based rejection. 

As discussed in the main text of the paper, removing instances using uncertainty estimates leads to higher performance as higher levels of uncertainty indicate the incorrectly predicted instances. 
Using epistemic uncertainty is more effective on Twitter 15 and Twitter 16 datasets, while aleatoric is better for the PHEME dataset. 
Softmax-based rejection also leads to improvements, but is outperformed by either aleatoric or epistemic estimates depending on the dataset.

\section{Per-fold unsupervised rejection.}
As we have explained in the experimental setup section of the main paper, during the cross-validation iterations each fold becomes a testing set once. We first aggregate predictions from each testing fold, and then perform evaluation and unsupervised rejection on the complete dataset. Alternatively, we could first perform the rejection procedure on each fold and then either aggregate the instances together for the evaluation (see tables \ref{perforldpheme}, \ref{perfoldtw15} and \ref{perfoldtw16}), or evaluate results on each fold separately (see table \ref{eachfoldpheme}). The outcomes are shown in tables \ref{perforldpheme}-\ref{perfoldtw16} below. 

The choice of set up does not affect the main conclusion of the paper regarding the benefits of using uncertainty estimates for this task. 
We chose to aggregate instances first because of the non-homogeneous sizes and label distributions of the folds in the PHEME dataset which introduces some artefacts. For example, Ebola-Essien event contains only 14 conversation threads, all of which are False rumours. This does not allow for meaningful conclusions about the model's performance, as it does not have all possible classes present. Furthermore when rejecting highly uncertain instances, the fold becomes even smaller.     

In table \ref{eachfoldpheme} we see drastic differences between folds in the PHEME dataset, which is not the case for the Twitter 15 and Twitter 16 datasets both of which contain folds balanced in size and label distribution. This also shows in the difference between the corresponding tables of the two set ups discussed in this section, which is more notable for PHEME (tables \ref{unsup} and \ref{perforldpheme}) than for the Twitter 15 (tables \ref{unsuptw15} and \ref{perfoldtw15}) and Twitter 16 (tables \ref{unsuptw16} and \ref{perfoldtw16}) datasets.

\section{Effect of Parameters on Uncertainty Estimates}
The methods we use for uncertainty estimates rely on a number of parameters.

For epistemic uncertainty the main parameter is the dropout probability as the method relies on applying dropout at testing time. Aleatoric uncertainty estimates depend on the number of times we perform sampling ($T$) and how much weight ($w$) the model places on optimising the loss function associated with uncertainty. 

\begin{figure*}[th]
   \subfloat[Aleatoric PHEME]{%
      \includegraphics[width=0.32\textwidth]{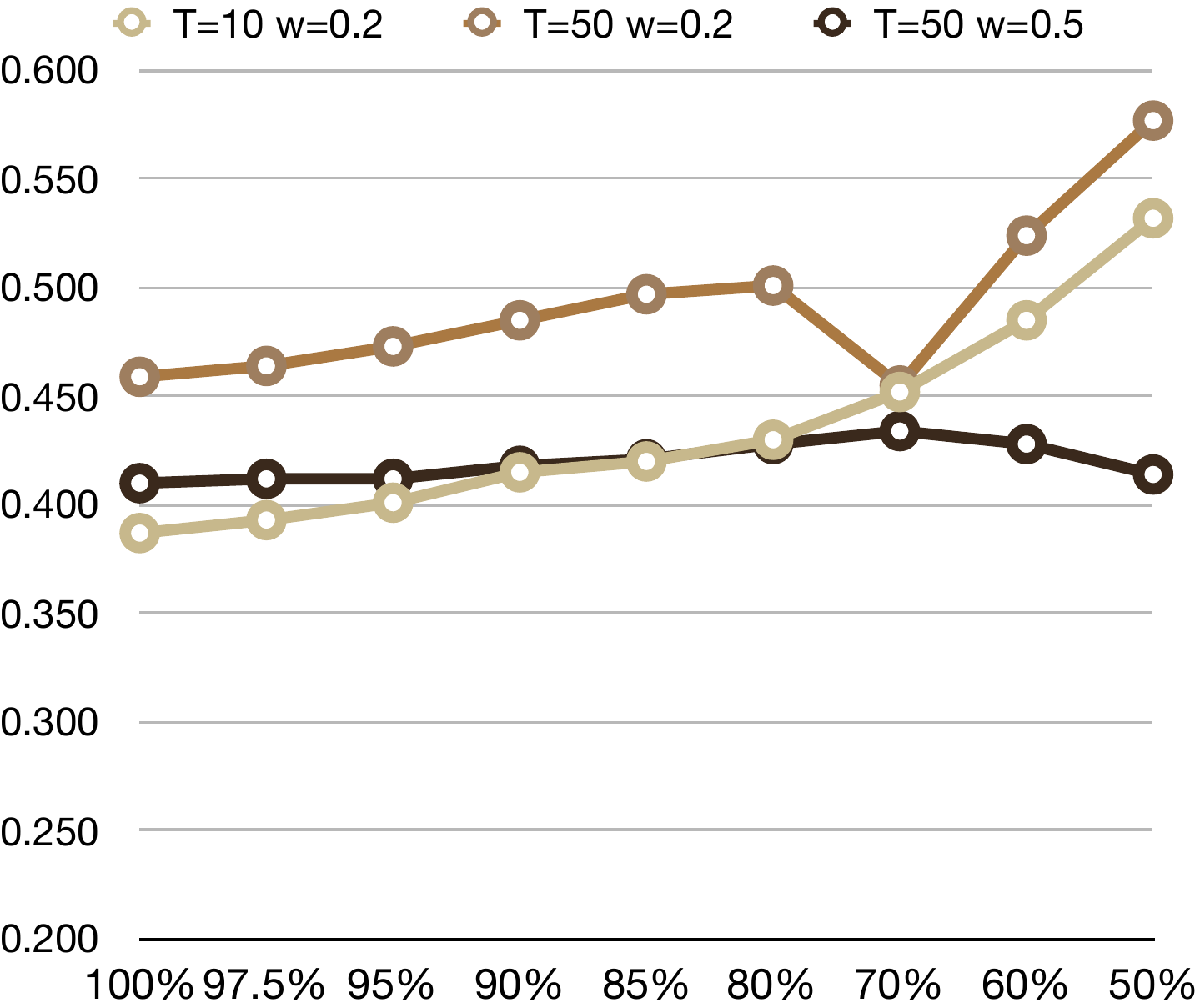}}
\hspace{\fill}
   \subfloat[Aleatoric Twitter 15]{%
     \includegraphics[width=0.32\textwidth]{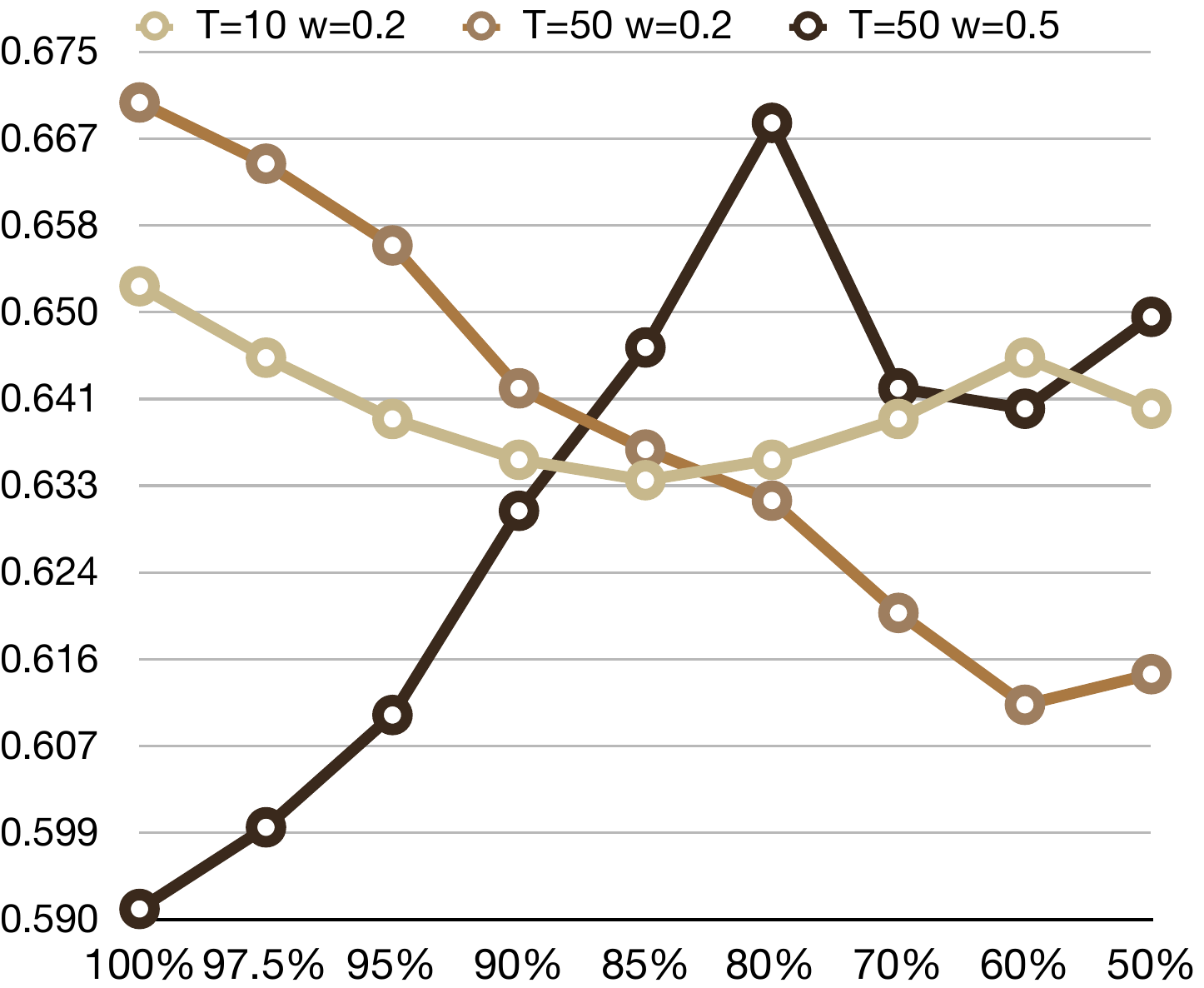}}
\hspace{\fill}
   \subfloat[Aleatoric Twitter 16]{%
      \includegraphics[width=0.32\textwidth]{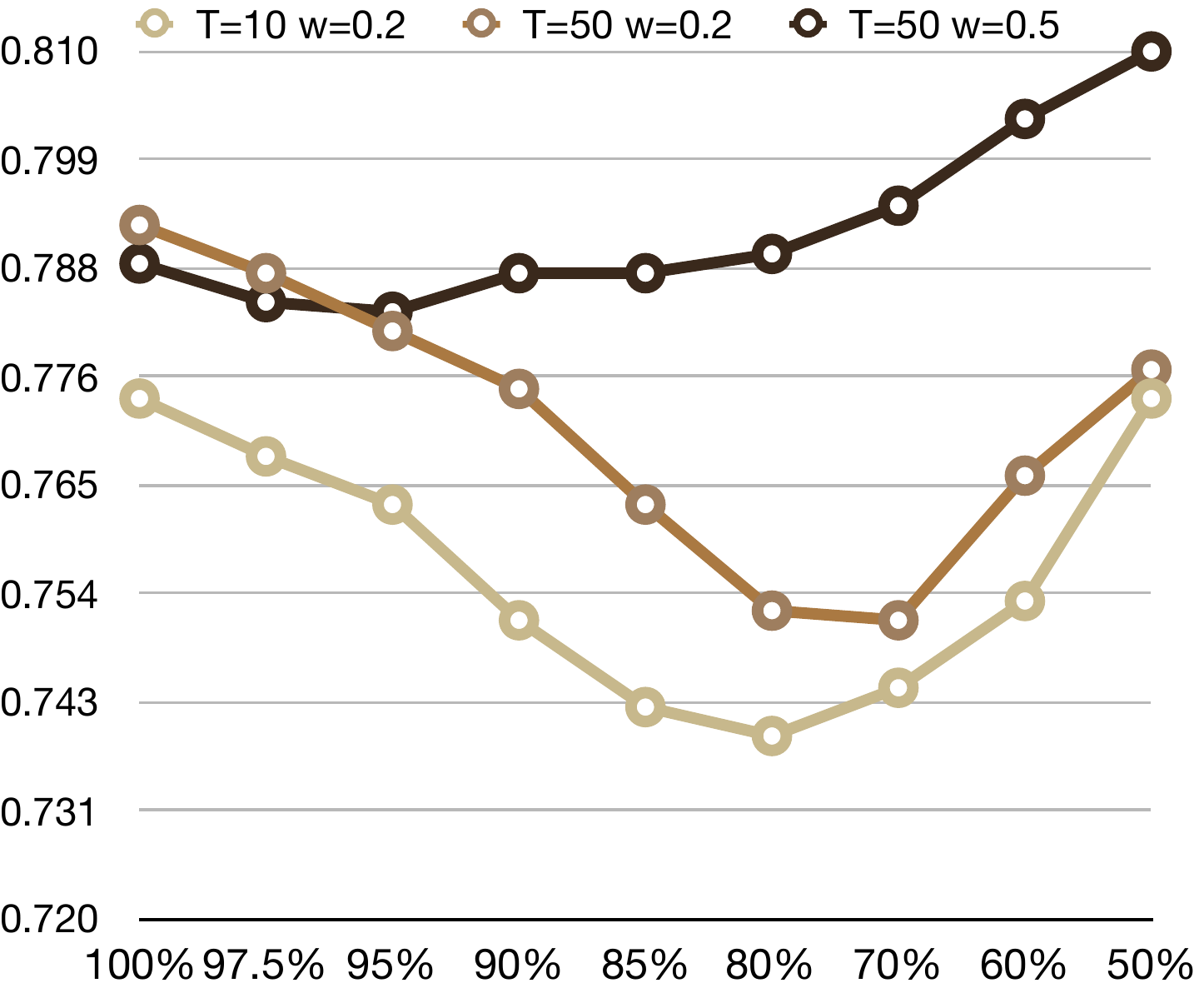}}\\
 \subfloat[Epistemic PHEME]{%
      \includegraphics[width=0.32\textwidth]{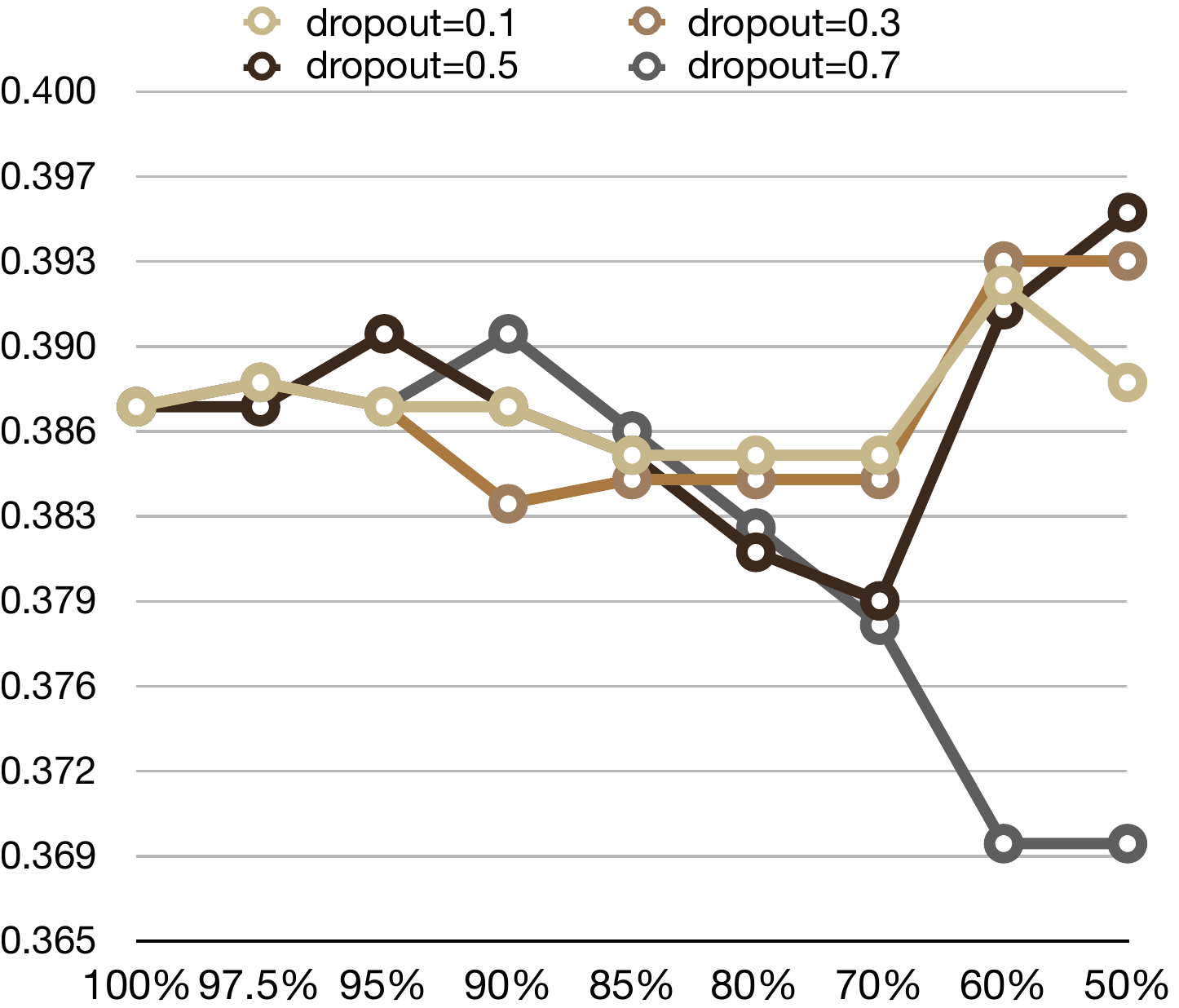}}
\hspace{\fill}
   \subfloat[Epistemic Twitter 15]{%
      \includegraphics[width=0.32\textwidth]{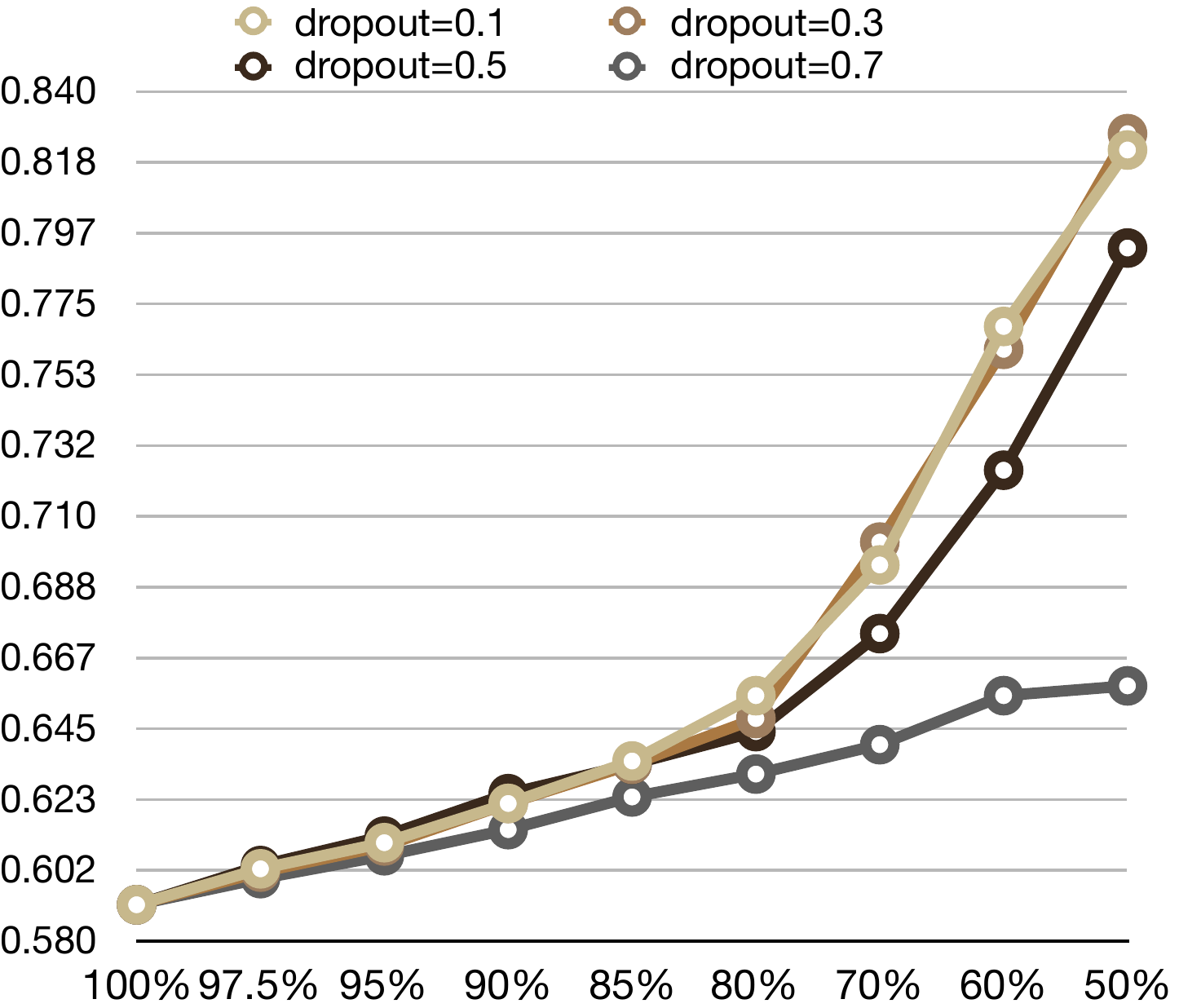}}
\hspace{\fill}
   \subfloat[Epistemic Twitter 16]{%
      \includegraphics[width=0.32\textwidth]{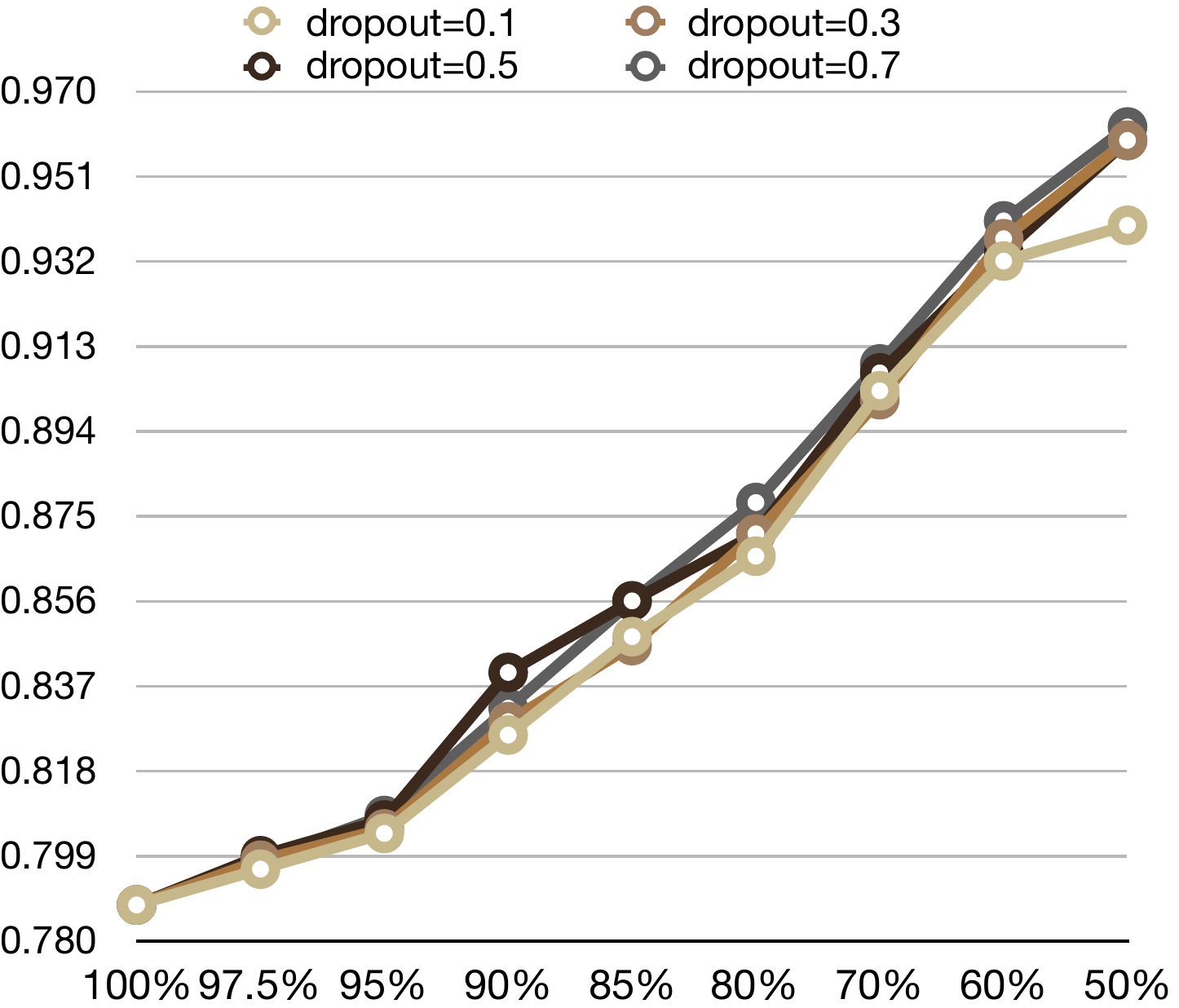}}\\     
\caption{Effect of parameters on uncertainty estimates.}
\label{fig_params}
\end{figure*}

We have performed a small parameter sweep comparing the output of models with testing dropout in [0.1, 0.3, 0.5, 0.7], T in [10, 50] and w in [0.2, 0.5]. Plots on Figure ~\ref{fig_params} show the effect of varying these parameters on unsupervised rejection outcomes in experiments on all datasets. In Figure ~\ref{fig_params} the Y-axis shows accuracy and the X-axis the proportion of the dataset on which it is measured. 

We see that the effect of parameters is dataset-dependent. The method for estimating aleatoric uncertainty affects a model's performance as it is incorporated in its loss function. By contrast estimating epistemic uncertainty using dropout at testing time does not have any effect on model performance.

On the plots for aleatoric uncertainty Figure~\ref{fig_params} (a-c) we see that changes in $T$ and $w$ strongly affect uncertainty estimates and the way they impact performance after unsupervised rejection.
On the balanced Twitter 15,16 datasets aleatoric uncertainty for low $T$ and $w$ values does not help disambiguate between correct and incorrect instances very well and needs to be tuned by increasing their values. However, that may lead to deterioration of model performance, introducing a trade-off. 

On the highly imbalanced PHEME dataset, aleatoric uncertainty estimates lead to improvements in performance for all parameter values, with the most increase observed when using a higher $T$ and $w=0.2$. We have not tested values of $T$ higher than 50, which could lead to further improvements. However it is likely there will be a maximum value after which we see no further improvements. 

Varying the dropout rate during testing leads to changes in epistemic uncertainty estimates and their effect on performance using unsupervised rejection (Figure~\ref{fig_params} (d-f)). The performance gains are observed for all three datasets. Increasing the dropout parameter from 0.1 to 0.3 in all datasets, and up to 0.5 in the PHEME and Twitter 16 datasets, leads to further improvements compared to lower values. However further increase of dropout to 0.7 starts to damage performance on the PHEME and Twitter 15 datasets.

\section{Uncertainty and Conversation Size}
\begin{figure*}[th]
   \subfloat[Aleatoric PHEME]{%
      \includegraphics[width=0.32\textwidth]{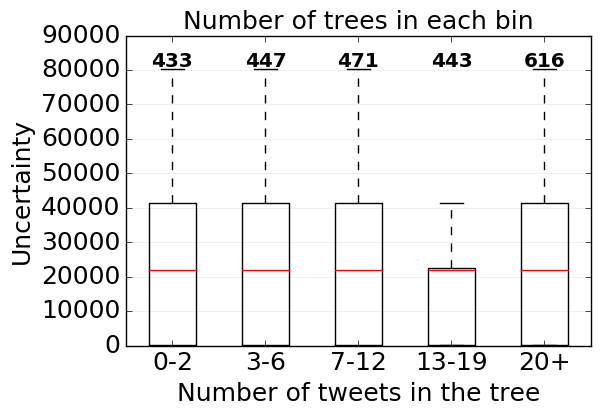}}
\hspace{\fill}
   \subfloat[Aleatoric Twitter 15]{%
     \includegraphics[width=0.32\textwidth]{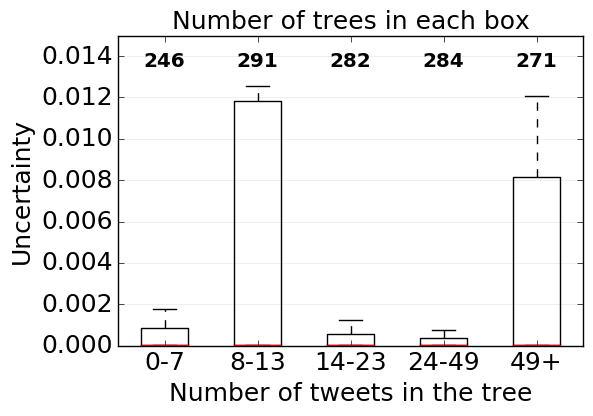}}
\hspace{\fill}
   \subfloat[Aleatoric Twitter 16]{%
      \includegraphics[width=0.32\textwidth]{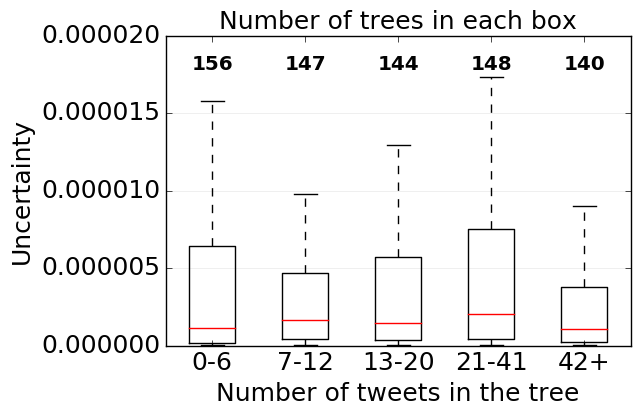}}\\
 \subfloat[Epistemic PHEME]{%
      \includegraphics[width=0.32\textwidth]{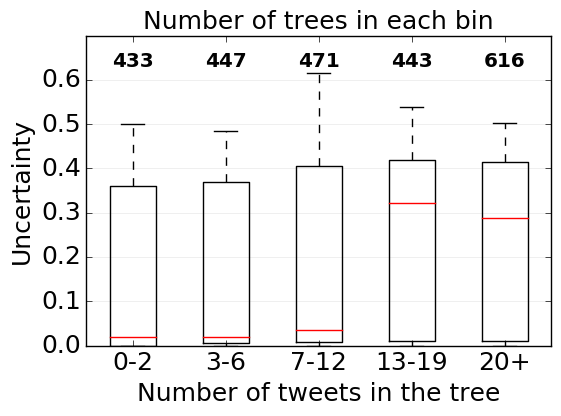}}
\hspace{\fill}
   \subfloat[Epistemic Twitter 15]{%
      \includegraphics[width=0.32\textwidth]{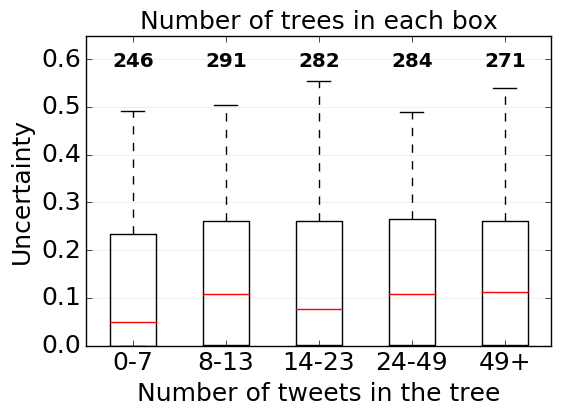}}
\hspace{\fill}
   \subfloat[Epistemic Twitter 16]{%
      \includegraphics[width=0.32\textwidth]{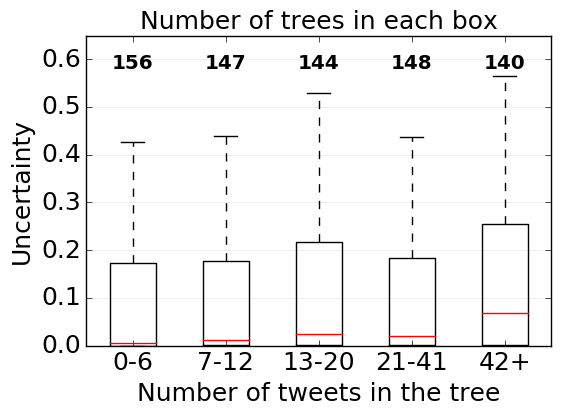}}\\ 
\subfloat[Softmax PHEME]{%
      \includegraphics[width=0.32\textwidth]{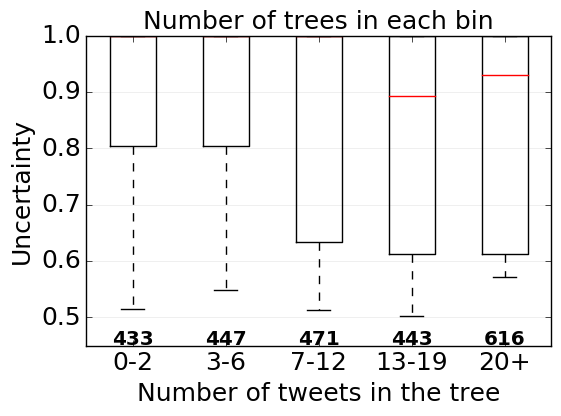}}
\hspace{\fill}
   \subfloat[Softmax Twitter 15]{%
      \includegraphics[width=0.32\textwidth]{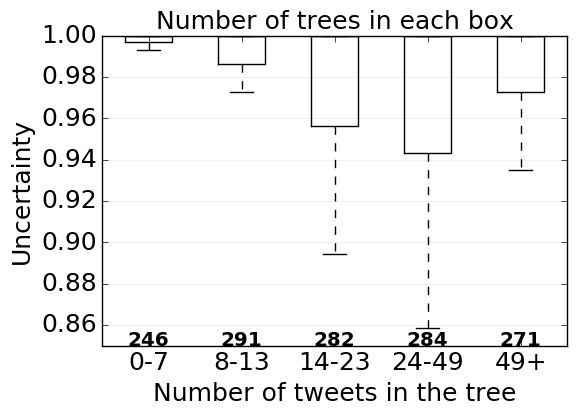}}
\hspace{\fill}
   \subfloat[Softmax Twitter 16]{%
      \includegraphics[width=0.32\textwidth]{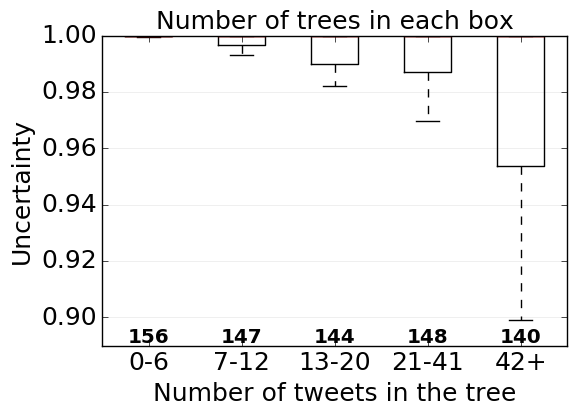}}\\ 
\caption{Boxplots showing uncertainty values grouped by the number of tweets in a conversation tree for 3 types of uncertainty estimates: aleatoric, epistemic, softmax. The Y-axis shows uncertainty (a-f) and confidence (g-i) values (a higher number indicates lower uncertainty). Numbers in bold show the number of conversations trees in each of the bins.}
\label{fig_length}
\end{figure*}

We have analysed how the size of the conversations affects uncertainty values. Figure \ref{fig_length} shows boxplots of uncertainty values of the conversations in all three datasets grouped by the number of tweets in each of them for aleatoric and epistemic uncertainty estimates as well as confidence levels (softmax). The conversations were grouped into equal sized bins, with resulting ranges of number of tweets are shown along the x-axis. 
We observe that for the confidence levels represented by the output of the softmax layer (Figure \ref{fig_length} (g,h,i)), conversations with a larger amount of tweets score lower values i.e., they have higher uncertainty. However for aleatoric and epistemic estimates (Figure \ref{fig_length} (a-f)) we do not observe a strong trend of uncertainty increase with the size of the conversation, so they seem to be more robust in this respect.
We have also performed this analysis using the number of branches in the conversation instead of the number of tweets and we have observed a similar pattern.

\section{Calibration}

\begin{figure*}[th]
   \subfloat[Aleatoric PHEME]{%
      \includegraphics[width=0.329\textwidth]{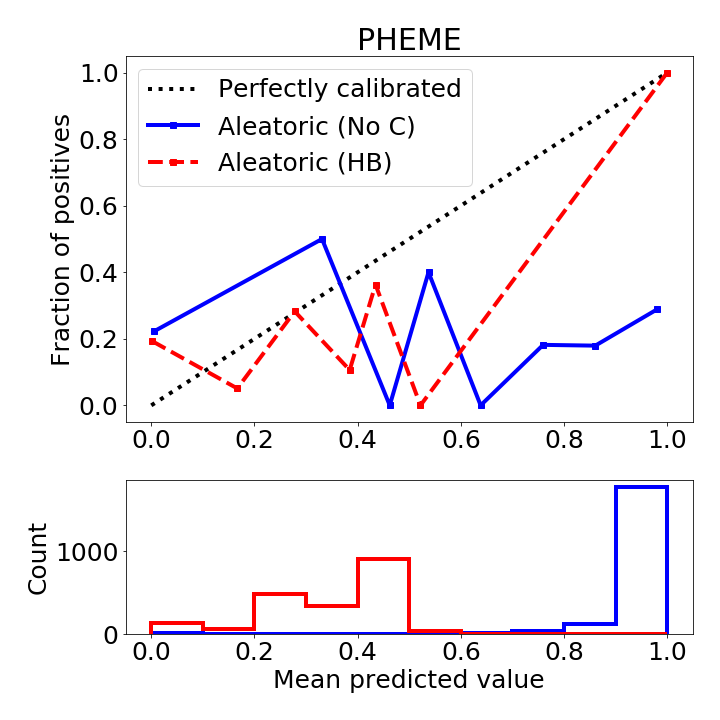}}
\hspace{\fill}
   \subfloat[Aleatoric Twitter 15]{%
     \includegraphics[width=0.329\textwidth]{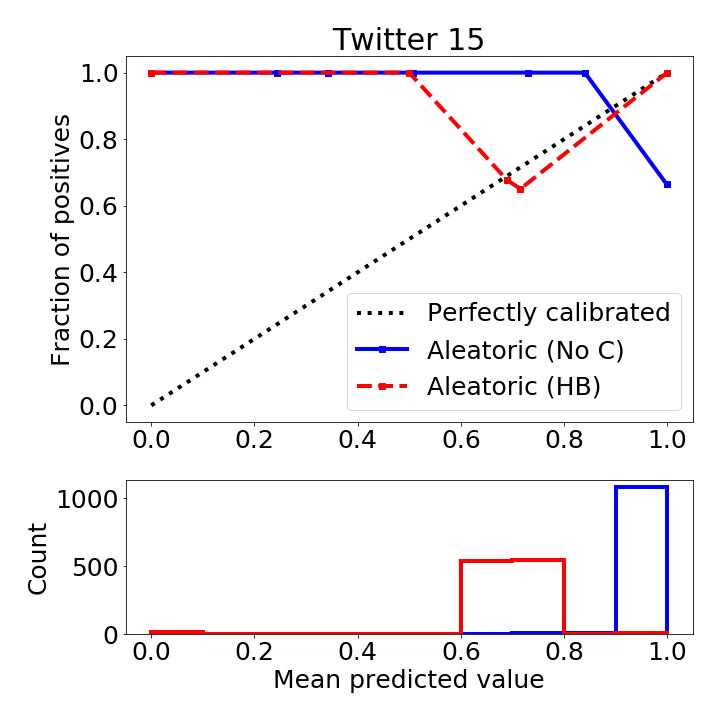}}
\hspace{\fill}
   \subfloat[Aleatoric Twitter 16]{%
      \includegraphics[width=0.329\textwidth]{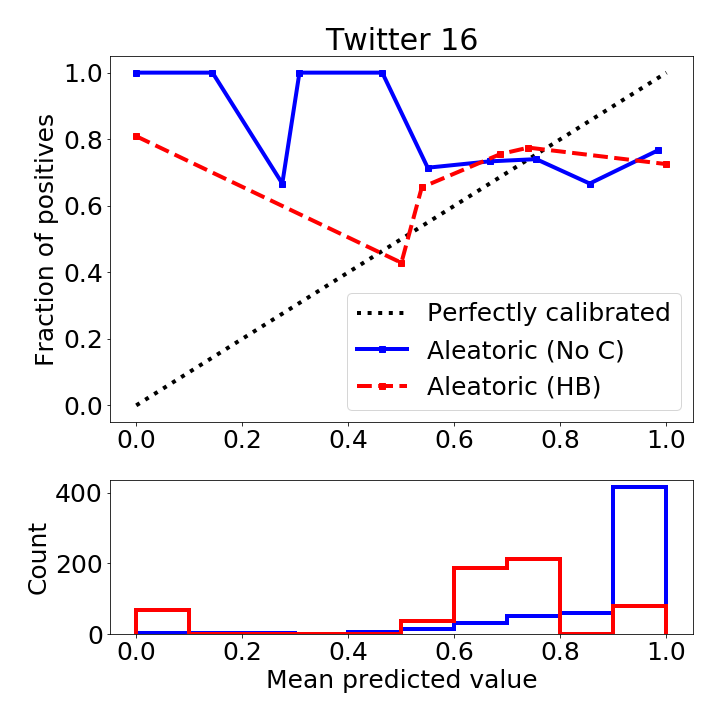}}\\
 \subfloat[Epistemic PHEME]{%
      \includegraphics[width=0.329\textwidth]{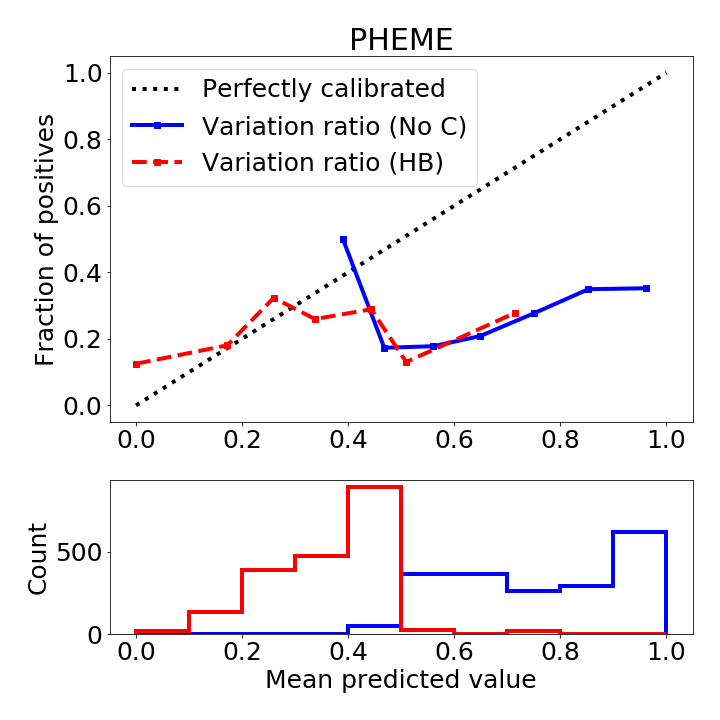}}
\hspace{\fill}
   \subfloat[Epistemic Twitter 15]{%
      \includegraphics[width=0.329\textwidth]{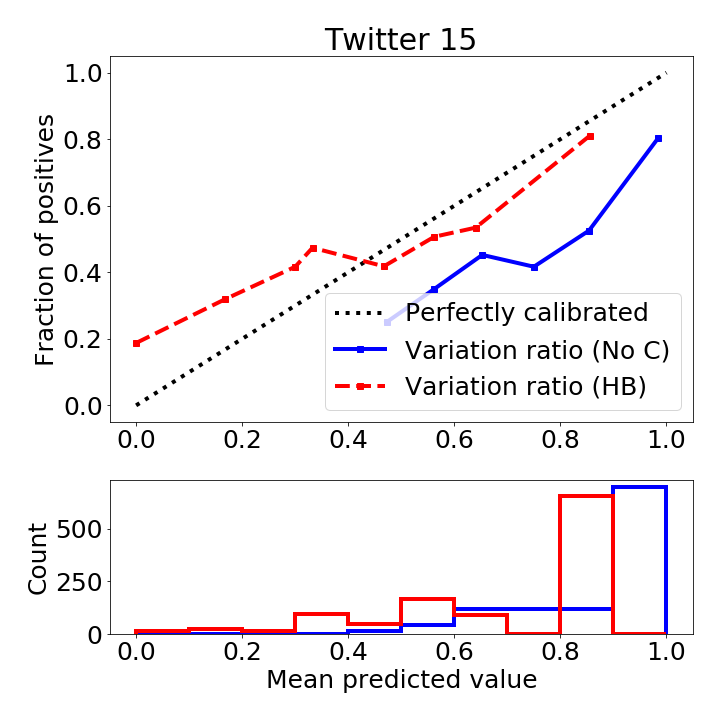}}
\hspace{\fill}
   \subfloat[Epistemic Twitter 16]{%
      \includegraphics[width=0.329\textwidth]{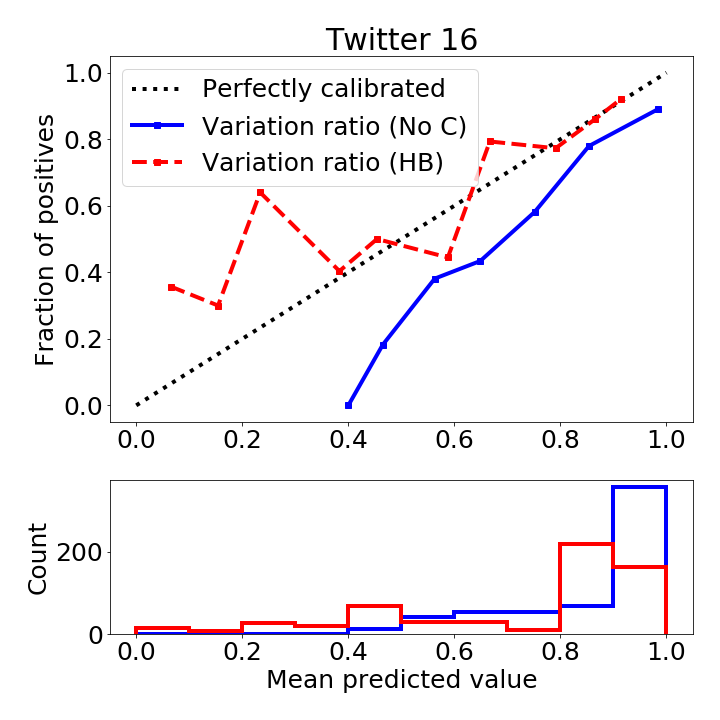}}\\ 
\subfloat[Softmax PHEME]{%
      \includegraphics[width=0.33\textwidth]{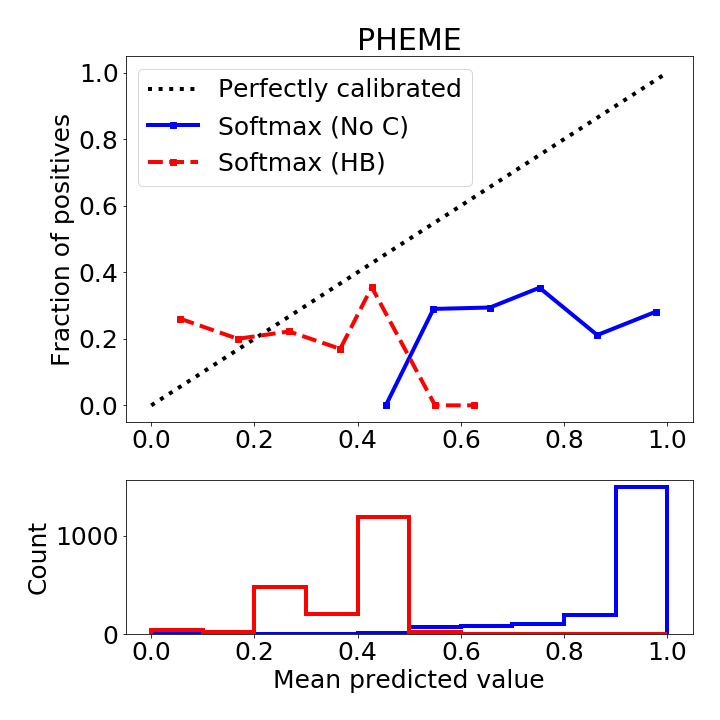}}
\hspace{\fill}
   \subfloat[Softmax Twitter 15]{%
      \includegraphics[width=0.329\textwidth]{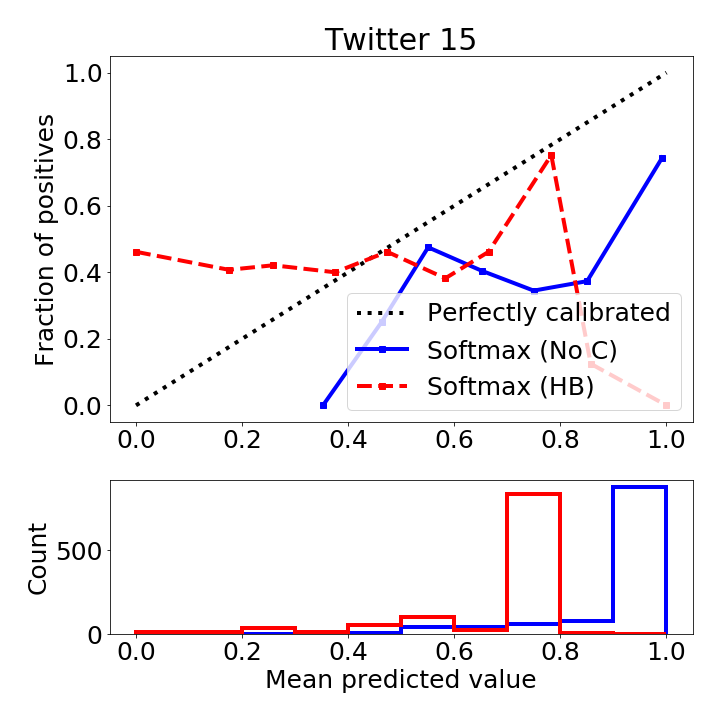}}
\hspace{\fill}
   \subfloat[Softmax Twitter 16]{%
      \includegraphics[width=0.329\textwidth]{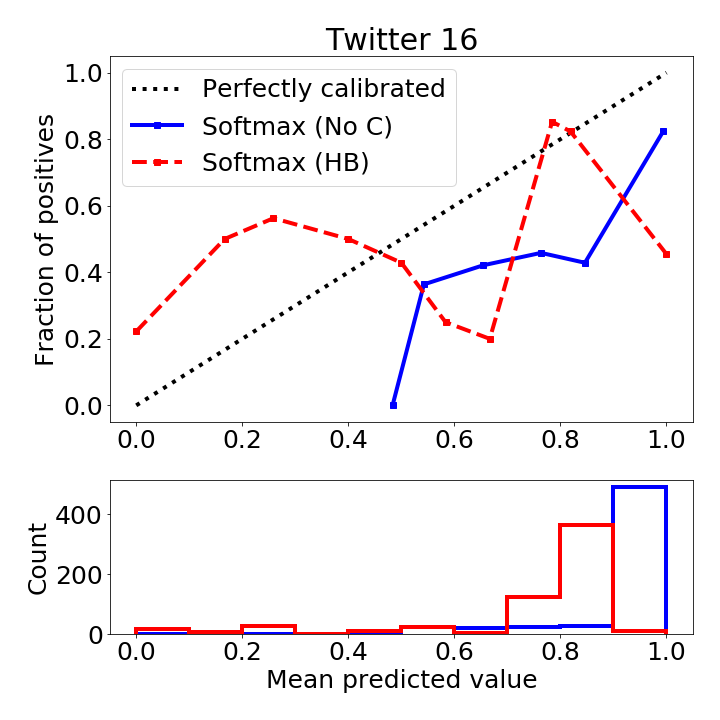}}\\ 
\caption{Reliability diagrams (calibration curves). X-axis shows confidence intervals, Y-axis shows accuracy at each interval (fraction of instances predicted correctly). Bottom plots show the number of instances in each interval. For both plots, blue - before calibration, red - after Histogram Binning.}
\label{fig_calibr}
\end{figure*}

Table \ref{tab:ece_supp} shows Expected Calibration Error (ECE) before and after the calibration process using the Histogram Binning method for all types of uncertainty. Figure~\ref{fig_calibr} shows corresponding reliability diagrams (calibration curves).
We use the experiment setup with one of the folds reserved as development set in order to train the calibration method. We convert uncertainty estimates $u$ into confidence scores as $1-u$, and for the aleatoric we normalise it to be in $[0,1]$. 
Calibration curves were plotted using the function from the scikit-learn package. Implementation of  ECE scores and Histogram Binning were adapted from \url{https://github.com/markus93/NN_calibration/blob/master/scripts/calibration/cal_methods.py}.

\section{Datasets}

Here we describe how to access the datasets used in the study. We use three publicly available datasets:

\subsection{PHEME}
The PHEME dataset can be downloaded here:\\
\url{https://figshare.com/articles/PHEME_dataset_for_Rumour_Detection_and_Veracity_Classification/6392078}
\subsection{Twitter 15,16}
The Twitter 15,16 datasets can be downloaded here:\\
\url{https://www.dropbox.com/s/7ewzdrbelpmrnxu/rumdetect2017.zip?dl=0}\\
It contains list of tweet ids belonging to the dataset. 

The split into folds for cross-validation is taken from here:
\url{https://github.com/majingCUHK/Rumor_RvNN/tree/master/nfold}\\

\begin{table*}[h]
\centering
\resizebox{\textwidth}{!}{%
\begin{tabular}{|l|l|l|l|l|l|l|l|l|l|l|}
\hline
\%     & \# removed & Random & Aleatoric                       & Entropy & Variance & Variation ratio & LCS   & MC    & RC    & E \\ \hline
100\%  & 0                    & 0.385  & 0.385                           & 0.385   & 0.385    & 0.385           & 0.385 & 0.385 & 0.385 & 0.385    \\ \hline
97.5\% & 60                   & 0.384  & 0.391                           & 0.388   & 0.387    & 0.386           & 0.386 & 0.387 & 0.387 & 0.385    \\ \hline
95\%   & 120                  & 0.384  & 0.397                           & 0.388   & 0.387    & 0.386           & 0.386 & 0.39  & 0.39  & 0.389    \\ \hline
90\%   & 240                  & 0.382  & 0.412                           & 0.385   & 0.387    & 0.387           & 0.387 & 0.389 & 0.389 & 0.389    \\ \hline
85\%   & 361                  & 0.384  & 0.417                           & 0.385   & 0.385    & 0.386           & 0.388 & 0.39  & 0.39  & 0.39     \\ \hline
80\%   & 481                  & 0.381  & 0.427                           & 0.385   & 0.385    & 0.387           & 0.388 & 0.387 & 0.387 & 0.387    \\ \hline
70\%   & 723                  & 0.374  & 0.448                           & 0.389   & 0.389    & 0.388           & 0.387 & 0.386 & 0.387 & 0.386    \\ \hline
60\%   & 964                  & 0.370  & 0.481                           & 0.387   & 0.396    & 0.394           & 0.389 & 0.377 & 0.377 & 0.376    \\ \hline
50\%   & 1205                 & 0.376  & \textbf{0.528} & 0.389   & 0.392    & 0.391           & 0.386 & 0.382 & 0.378 & 0.381    \\ \hline
\end{tabular}%
}
\caption{Performance (accuracy) after unsupervised rejection on PHEME dataset for all types of uncertainty. LCS -- Least Confidence Sampling; MC -- Margin of Confidence, RC -- Ratio of Confidence and E -- Entropy based on a single output  of a softmax layer (as opposed to Entropy, Variance and Variation ratio that are based on multiple softmax samples).}
\label{unsup}
\end{table*}
\begin{table*}[h]
\centering
\resizebox{\textwidth}{!}{%
\begin{tabular}{|l|l|l|l|l|l|l|l|l|l|l|}
\hline
\%      & \# removed & Random & Aleatoric & Entropy & Variance & Variation ratio & LCS   & MC    & RC    & E     \\ \hline
100.0\% & 0          & 0.591  & 0.591     & 0.591   & 0.591    & 0.591           & 0.591 & 0.591 & 0.591 & 0.591 \\ \hline
97.5\%  & 34         & 0.589  & 0.599     & 0.603   & 0.599    & 0.602           & 0.601 & 0.601 & 0.601 & 0.6   \\ \hline
95.0\%  & 68         & 0.593  & 0.61      & 0.609   & 0.609    & 0.609           & 0.609 & 0.609 & 0.609 & 0.609 \\ \hline
90.0\%  & 137        & 0.592  & 0.63      & 0.625   & 0.627    & 0.622           & 0.621 & 0.621 & 0.620 & 0.622 \\ \hline
85.0\%  & 206        & 0.597  & 0.647     & 0.637   & 0.646    & 0.634           & 0.634 & 0.634 & 0.631 & 0.634 \\ \hline
80.0\%  & 274        & 0.599  & 0.668     & 0.648   & 0.665    & 0.657           & 0.630 & 0.631 & 0.633 & 0.630 \\ \hline
70.0\%  & 412        & 0.577  & 0.642     & 0.669   & 0.718    & 0.699           & 0.660 & 0.661 & 0.660 & 0.660 \\ \hline
60.0\%  & 549        & 0.596  & 0.64      & 0.679   & 0.77     & 0.765           & 0.684 & 0.684 & 0.684 & 0.684 \\ \hline
50.0\%  & 687        & 0.598  & 0.649     & 0.677   & 0.817    & \textbf{0.821}  & 0.723 & 0.722 & 0.721 & 0.723 \\ \hline
\end{tabular}%
}
\caption{Performance (accuracy) after unsupervised rejection on Twitter 15 dataset for all types of uncertainty. LCS -- Least Confidence Sampling; MC -- Margin of Confidence, RC -- Ratio of Confidence and E -- Entropy based on a single output  of a softmax layer.}
\label{unsuptw15}
\end{table*}
\begin{table*}[h!]
\centering
\resizebox{\textwidth}{!}{%
\begin{tabular}{|l|l|l|l|l|l|l|l|l|l|l|}
\hline
\%      & \# removed & Random & Aleatoric & Entropy & Variance       & Variation ratio & LCS   & MC    & RC    & E     \\ \hline
100.0\% & 0          & 0.788  & 0.788     & 0.788   & 0.788          & 0.788           & 0.788 & 0.788 & 0.788 & 0.788 \\ \hline
97.5\%  & 18         & 0.789  & 0.784     & 0.798   & 0.794          & 0.796           & 0.795 & 0.795 & 0.795 & 0.796 \\ \hline
95.0\%  & 36         & 0.787  & 0.783     & 0.808   & 0.805          & 0.805           & 0.800 & 0.801 & 0.800 & 0.804 \\ \hline
90.0\%  & 73         & 0.787  & 0.787     & 0.837   & 0.828          & 0.829           & 0.828 & 0.828 & 0.826 & 0.828 \\ \hline
85.0\%  & 110        & 0.786  & 0.787     & 0.856   & 0.85           & 0.856           & 0.848 & 0.848 & 0.848 & 0.848 \\ \hline
80.0\%  & 146        & 0.789  & 0.789     & 0.881   & 0.868          & 0.869           & 0.864 & 0.864 & 0.862 & 0.866 \\ \hline
70.0\%  & 220        & 0.794  & 0.794     & 0.905   & 0.907          & 0.901           & 0.905 & 0.905 & 0.905 & 0.905 \\ \hline
60.0\%  & 294        & 0.787  & 0.803     & 0.939   & 0.937          & 0.937           & 0.925 & 0.925 & 0.925 & 0.925 \\ \hline
50.0\%  & 367        & 0.78   & 0.81      & 0.954   & \textbf{0.957} & \textbf{0.957}  & 0.951 & 0.951 & 0.951 & 0.954 \\ \hline
\end{tabular}%
}
\caption{Performance (accuracy) after unsupervised rejection on Twitter 16 dataset for all types of uncertainty. LCS -- Least Confidence Sampling; MC -- Margin of Confidence, RC -- Ratio of Confidence and E -- Entropy based on a single output  of a softmax layer.}
\label{unsuptw16}
\end{table*}
\begin{table*}[]
\centering
\begin{tabular}{|l|l|l|l|l|l|l|l|l|l|}
\hline
\%    & EE & FE & GU & OT & PT & PM   & SS    & CH   & GW \\ \hline
100\% & 0.429        & 0.062    & 0.459   & 0.589          & 0.240           & 0.325          & 0.588          & 0.353          & 0.139             \\ \hline
90\%  & 0.385        & 0.053    & 0.491   & 0.601          & 0.217          & 0.333          & 0.600            & 0.360           & 0.140              \\ \hline
80\%  & 0.417        & 0.030     & 0.469   & 0.615          & 0.196          & 0.356          & 0.608          & 0.370           & 0.131             \\ \hline
70\%  & 0.400          & 0.02     & 0.442   & 0.62           & 0.186          & 0.337          & 0.628          & 0.376          & 0.114             \\ \hline
60\%  & 0.333        & 0.023    & 0.378   & 0.638          & 0.181          & 0.342          & 0.653          & 0.391          & 0.112             \\ \hline
50\%  & 0.429        & 0.014    & 0.387   & \textbf{0.655} & 0.174          & \textbf{0.333} & \textbf{0.674} & \textbf{0.422} & 0.118             \\ \hline
\end{tabular}%
\caption{Unsupervised rejection using variation ratio uncertainty estimates for each event--fold in the PHEME dataset. EE -- Ebola-Essien; FE -- Ferguson unrest; GU -- Gurlitt; OT -- Ottawa shooting; PT -- Prince-Toronto; PM -- Putin missing; SS -- Sydney Siege; CH -- Charlie Hebdo; GW -- Germanwings crash.}
\label{eachfoldpheme}
\end{table*}
\begin{table*}[]
\centering
\begin{tabular}{|l|l|l|l|l|l|l|l|l|}
\hline
\%    & Aleatoric & Variation ratio & Entropy & Variance & LCS   & MC    & RC    & E     \\ \hline
100\% & 0.385     & 0.385           & 0.385   & 0.385    & 0.385 & 0.385 & 0.385 & 0.385 \\ \hline
90\%  & 0.395     & 0.389           & 0.388   & 0.389    & 0.392 & 0.392 & 0.394 & 0.391 \\ \hline
80\%  & 0.397     & 0.390           & 0.393   & 0.390     & 0.392 & 0.392 & 0.392 & 0.392 \\ \hline
70\%  & 0.400     & 0.391           & 0.393   & 0.391    & 0.399 & 0.398 & 0.395 & 0.399 \\ \hline
60\%  & 0.393     & 0.401           & 0.399   & 0.405    & 0.399 & 0.399 & 0.396 & 0.399 \\ \hline
50\%  & 0.386     & \textbf{0.413}           & \textbf{0.413}   & \textbf{0.413}    & 0.400   & 0.401 & 0.395 & 0.400   \\ \hline
\end{tabular}%
\caption{Performance (accuracy) after per-fold unsupervised rejection on PHEME dataset for all types of uncertainty. LCS -- Least Confidence Sampling; MC -- Margin of Confidence, RC -- Ratio of Confidence and E -- Entropy based on a single output of a softmax layer (as opposed to Entropy, Variance and Variation ratio that are based on multiple softmax samples).}
\label{perforldpheme}
\end{table*}
\begin{table*}[]
\centering
\begin{tabular}{|l|l|l|l|l|l|l|l|l|}
\hline
\%    & Aleatoric & Variation ratio & Entropy & Variance       & LCS   & MC    & RC    & E     \\ \hline
100\% & 0.591     & 0.591           & 0.591   & 0.591          & 0.591 & 0.591 & 0.591 & 0.591 \\ \hline
90\%  & 0.566     & 0.625           & 0.625   & 0.622          & 0.619 & 0.619 & 0.619 & 0.623 \\ \hline
80\%  & 0.558     & 0.650            & 0.657   & 0.652          & 0.654 & 0.652 & 0.651 & 0.653 \\ \hline
70\%  & 0.569     & 0.699           & 0.697   & 0.699          & 0.675 & 0.675 & 0.673 & 0.674 \\ \hline
60\%  & 0.596     & 0.725           & 0.714   & 0.724          & 0.707 & 0.708 & 0.708 & 0.707 \\ \hline
50\%  & 0.603     & 0.753           & 0.753   & \textbf{0.756} & 0.741 & 0.741 & 0.741 & 0.741 \\ \hline
\end{tabular}%
\caption{Performance (accuracy) after per-fold unsupervised rejection on Twitter 15 dataset for all types of uncertainty. LCS -- Least Confidence Sampling; MC -- Margin of Confidence, RC -- Ratio of Confidence and E -- Entropy based on a single output of a softmax layer (as opposed to Entropy, Variance and Variation ratio that are based on multiple softmax samples).}
\label{perfoldtw15}
\end{table*}

\begin{table*}[]
\centering
\begin{tabular}{|l|l|l|l|l|l|l|l|l|}
\hline
\%    & Aleatoric & Variation ratio & Entropy        & Variance & LCS   & MC    & RC    & E     \\ \hline
100\% & 0.788     & 0.788           & 0.788          & 0.788    & 0.788 & 0.788 & 0.788 & 0.788 \\ \hline
90\%  & 0.783     & 0.830            & 0.833          & 0.821    & 0.816 & 0.818 & 0.821 & 0.816 \\ \hline
80\%  & 0.782     & 0.870            & 0.873          & 0.870     & 0.866 & 0.866 & 0.866 & 0.865 \\ \hline
70\%  & 0.784     & 0.898           & 0.902          & 0.903    & 0.902 & 0.902 & 0.900   & 0.902 \\ \hline
60\%  & 0.810      & 0.928           & 0.934          & 0.932    & 0.921 & 0.921 & 0.921 & 0.921 \\ \hline
50\%  & 0.835     & 0.954           & \textbf{0.962} & 0.957    & 0.938 & 0.940  & 0.949 & 0.949 \\ \hline
\end{tabular}%
\caption{Performance (accuracy) after per-fold unsupervised rejection on Twitter 15 dataset for all types of uncertainty. LCS -- Least Confidence Sampling; MC -- Margin of Confidence, RC -- Ratio of Confidence and E -- Entropy based on a single output of a softmax layer (as opposed to Entropy, Variance and Variation ratio that are based on multiple softmax samples).}
\label{perfoldtw16}
\end{table*}

\begin{table*}[]
\centering
\begin{tabular}{|l|l|l|l|l|l|l|l|l|l|l|}
\hline
\multirow{2}{*}{} & \multicolumn{5}{l|}{No calibration}   & \multicolumn{5}{l|}{Histogram Binning} \\ \cline{2-11} 
                  & S     & A     & VR    & E     & VAR   & S      & A     & VR    & E     & VAR   \\ \hline
PHEME             & 0.646 & 0.683 & 0.492 & 0.292 & 0.295 & 0.173  & 0.088 & 0.111 & 0.119 & 0.108 \\ \hline
Twitter 15        & 0.265 & 0.333 & 0.216 & 0.119 & 0.144 & 0.056  & 0.039 & 0.062 & 0.065 & 0.066 \\ \hline
Twitter 16        & 0.191 & 0.196 & 0.121 & 0.080 & 0.109 & 0.164  & 0.079 & 0.044 & 0.058 & 0.056 \\ \hline
\end{tabular}%
\caption{Expected Calibration Error before and after applying calibration over uncertainty estimates. S - softmax (LCS), A - aleatoric uncertainty, VR - variation ratio, E - entropy, VAR - variance.}
\label{tab:ece_supp}
\end{table*}

\end{document}